\documentclass[10pt,twocolumn,letterpaper]{article}

\usepackage[final]{cvpr}              % To produce the CAMERA-READY version
\usepackage{dsfont}
\usepackage[utf8]{inputenc} % allow utf-8 input
\usepackage[T1]{fontenc}    % use 8-bit T1 fonts
\usepackage{url}            % simple URL typesetting
\usepackage{booktabs}       % professional-quality tables
\usepackage{amsfonts}       % blackboard math symbols
\usepackage{nicefrac}       % compact symbols for 1/2, etc.
\usepackage{microtype}      % microtypography

\usepackage{amsmath}
\usepackage{amssymb}
\usepackage{listings}
\usepackage{caption}
\usepackage{bbm}
\usepackage{arydshln}

\usepackage{listings}
\usepackage{placeins}
\usepackage{nicefrac}
\usepackage{xurl}
\usepackage{enumitem}
\usepackage{tabularx}
\usepackage{nicefrac}
\usepackage{makecell}
\usepackage{xspace}
\usepackage{graphbox}
\usepackage{pifont}
\usepackage{enumitem}
\usepackage{colortbl}
\usepackage{multirow}
\usepackage{diagbox}
\usepackage{array}
\usepackage{wrapfig}
\usepackage{subcaption}
\usepackage{footmisc}
\usepackage{color}
\usepackage{stfloats}
\usepackage{slashbox}
\usepackage{diagbox}
\usepackage{float}
\usepackage{mathrsfs}
\usepackage{apptools}
\usepackage{bbm}
\usepackage{soul}

\graphicspath{ {figures} }

\usepackage[dvipsnames]{xcolor}
\usepackage{tcolorbox}
\usepackage{url}
\definecolor{cvprblue}{rgb}{0.21,0.49,0.74}

\usepackage[pagebackref,breaklinks,colorlinks,allcolors=cvprblue]{hyperref}

\def\paperID{*****} % *** Enter the Paper ID here
\def\confName{CVPR}
\def\confYear{2026}

\newcolumntype{C}[1]{>{\centering\arraybackslash}p{#1}}
\newcolumntype{L}[1]{>{\raggedright\arraybackslash}p{#1}}
\newcolumntype{R}[1]{>{\raggedleft\arraybackslash}p{#1}}
\newlength\newl
\newlength\newlc

\newlength\colwidth
\newlength\figwidth

\newcommand{\clevrer}{\textsc{Clevrer}\xspace}
\newcommand{\videoespresso}{\textsc{VideoEspresso}\xspace}
\newcommand{\videomme}{\textsc{Video-MME}\xspace}
\newcommand{\vidhal}{\textsc{VidHal}\xspace}
\newcommand{\eventhallusion}{\textsc{EventHallusion}\xspace}
\newcommand{\eventhallusionshort}{\textsc{EventHall}\xspace}
\newcommand{\vsi}{\textsc{VSI-Bench}\xspace}
\newcommand{\mvbench}{\textsc{MVBench}\xspace}
\newcommand{\nextqa}{\textsc{NextQA}\xspace}
\newcommand{\cofdata}{\textsc{CoF-Data}\xspace}
\newcommand{\cofdatareal}{\cofdata{}\textsubscript{real}\xspace}
\newcommand{\cofdatasyn}{\cofdata{}\textsubscript{synth}\xspace}
\newcommand{\ivls}{InternVL2.5-4B\xspace}
\newcommand{\ivlb}{InternVL3-8B\xspace}
\newcommand{\phiv}{Phi-3.5-Vision-4B\xspace}

\newcommand{\cofb}{CoF-8B\xspace}

\newlength\myindent

\definecolor{brightpurple}{RGB}{160,0,255}

\definecolor{lightorange}{RGB}{255, 229, 180}
\definecolor{lightblue}{RGB}{155, 155, 255}
\definecolor{lightgreen}{RGB}{182, 235, 185}

\newcommand{\myparagraph}[1]{
\textbf{#1}
}
\usepackage{amsmath,amsfonts,bm}

\def\eqref#1{Eq.~(\ref{#1})}
\def\1{\bm{1}}

\DeclareMathAlphabet{\mathsfit}{\encodingdefault}{\sfdefault}{m}{sl}
\SetMathAlphabet{\mathsfit}{bold}{\encodingdefault}{\sfdefault}{bx}{n}

\tcbuselibrary{listingsutf8}
\newtcblisting{promptbox}[1][]{
  colback=gray!10,
  colframe=gray!80,
  listing only,
  listing options={
    basicstyle=\ttfamily\small,
    breaklines=true,
    breakatwhitespace=true
  },
  title={#1},
  boxrule=0.5pt,
  arc=2mm,
  width=\linewidth,     % Ensures it fits the page
  left=1mm, right=1mm,  % Padding inside box
  top=1mm, bottom=1mm
}
\newcommand{\upblue}[1]{\textbf{\textcolor{blue}{\text{ #1$\uparrow$}}}}

\title{Chain-of-Frames: Advancing Video Understanding in Multimodal LLMs\\ via Frame-Aware Reasoning
}

\author{Sara Ghazanfari$^{1*}$
\qquad Francesco Croce$^{2}$
\qquad Nicolas Flammarion$^{2}$ \vspace{0.05cm}\\
\qquad Prashanth Krishnamurthy$^{1}$ 
\qquad Farshad Khorrami$^{1}$
\qquad Siddharth Garg$^{1}$ 
\vspace{0.2cm} \\ 
$^{1}$New York University, US\; $^{2}$EPFL, Switzerland\; 
$^{*}$Correspondence: \texttt{sg7457@nyu.edu}
\vspace{-0.2cm}
}

\begin{document}
\twocolumn[{
\maketitle

\begin{center}

  \begin{minipage}[b]{0.63\linewidth}
    \centering
    \includegraphics[width=\linewidth, trim=50mm 0mm 0mm 0mm, clip]{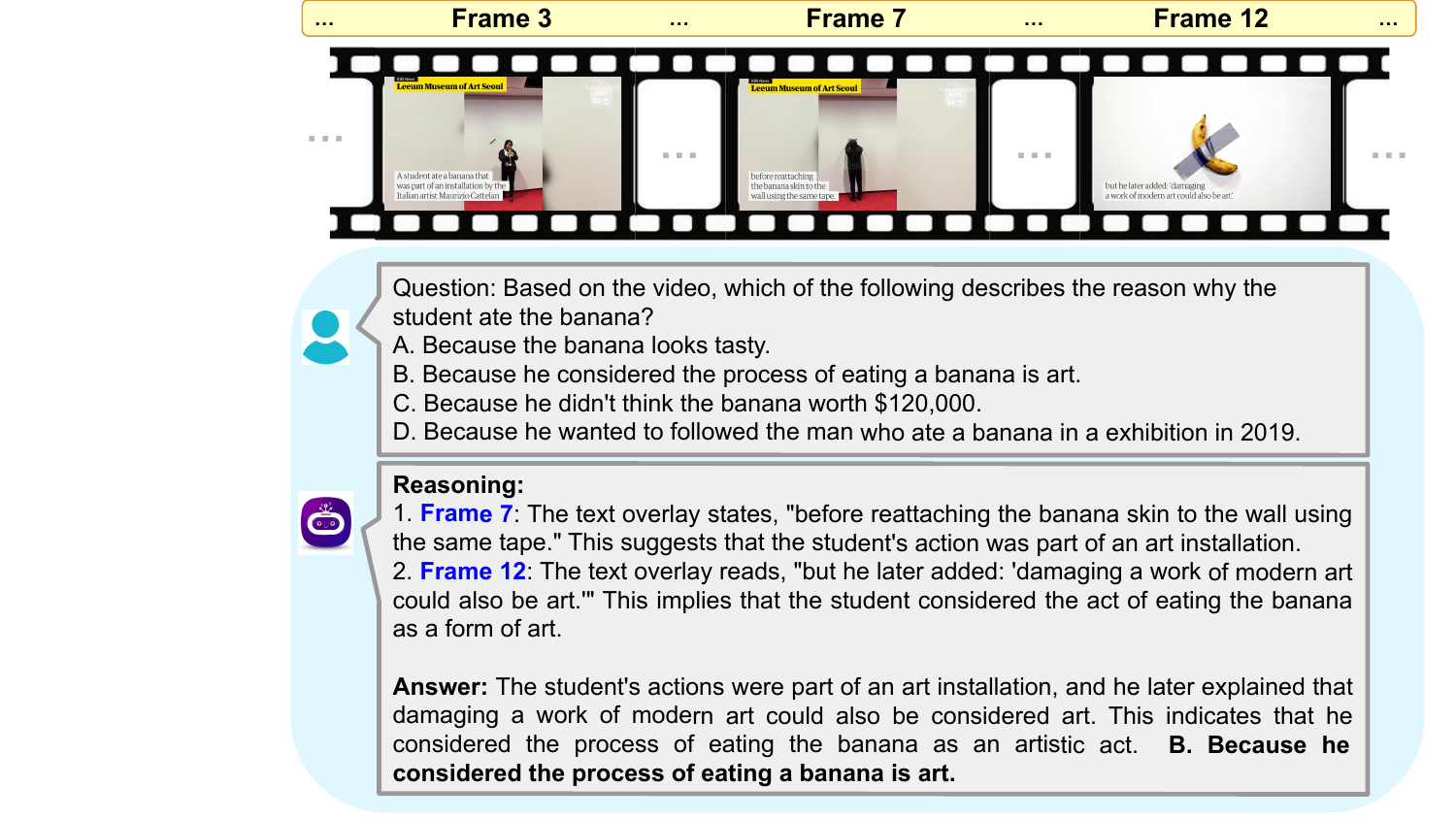}
    \begin{center}
    {\small \textbf{(a)} A Chain-of-Frames reasoning trace generated by our CoF-4B model.}
\end{center}
    % \captionof{figure}{\textbf{(a)} A Chain-of-Frames reasoning trace generated by our CoF-4B model.}
    \label{fig:teaser-exp}
  \end{minipage}
  \hfill
  \begin{minipage}[b]{0.35\linewidth}
    \centering
    \includegraphics[width=\linewidth, trim=0mm 0mm 0mm 0mm, clip]{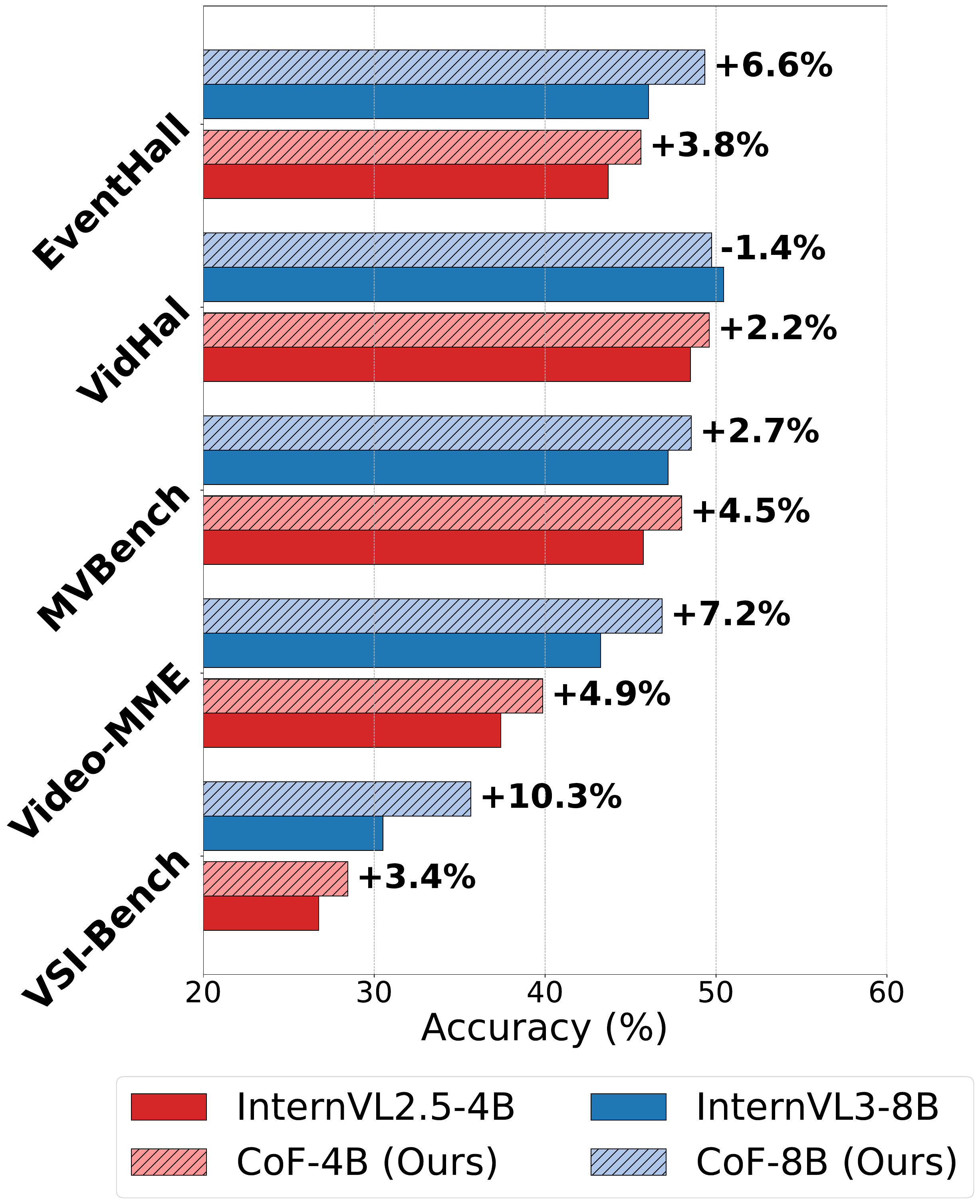}
    \begin{center}
    {\small \textbf{(b)} CoF models vs the baseline models.}
\end{center}
    % \captionof{figure}{\textbf{(b)} CoF models vs the baseline models.}
    \label{fig:teaser-acc}
  \end{minipage}
    \vspace{-2mm}
  \captionof{figure}{\textbf{(a)} Chain-of-Frames (CoF) reasoning generated by our CoF-\ivls model, referencing the key frames 
    to answer the question (from \videomme).
    \textbf{(b)}
    Comparison of accuracy across multiple video understanding benchmarks between baseline models (\ivls and \ivlb) and their CoF-enhanced counterparts: our models outperform the baselines in nearly every case.
  }
  \label{fig:teaser}
\end{center}
}]

\begin{abstract}
Recent work has shown that eliciting Large Language Models (LLMs) to generate reasoning traces in natural language before answering the user's request can significantly improve their performance across tasks.
This approach has been extended to multimodal LLMs, where the models can produce chains-of-thoughts (CoT) about the content of input images and videos. For video inputs, prior works use complex multi-step pipelines that extract and include relevant frames from videos in the CoT, or produce simpler single-stage reasoning traces at the expense of poor temporal grounding.
%have failed to produce CoTs with 
Here, we propose the first video LLMs with single-stage reasoning that includes explicit references to relevant frames, thereby reducing temporal inconsistencies in the reasoning process. Our approach is simple, unified, and self-contained, employing a single-stage inference to handle complex video understanding tasks without relying on auxiliary modules for frame selection or caption generation.
For this, we first create \cofdata, a large dataset of diverse questions, answers, and corresponding frame-grounded reasoning traces from both natural and synthetic videos, spanning various topics and tasks.
Our models, obtained fine-tuning video LLMs on this chain-of-frames (CoF) data, generate reasoning traces that accurately identify key frames to answer given questions.
In turn, this consistently improves performance across multiple video understanding benchmarks. 
Surprisingly, we find that synthetic data alone, despite being out-of-distribution with respect to these real-world benchmarks, provides a significant boost in model accuracy. Code available at \href{https://github.com/SaraGhazanfari/CoF}{GitHub}.
\end{abstract}
\vspace{-4mm}

\section{Introduction}
Large Language Models (LLMs) are able to perform step-by-step reasoning, widely known as chain-of-thoughts (CoT) \citep{wei2022chain, kojima2022large}.
This capability has been implicitly integrated into state-of-the-art systems such as OpenAI’s o1/o3 models~\citep{openai_o1o3_2024} and DeepSeek R1~\citep{guo2025deepseek}, 
contributing to their remarkable performance and improving the interpretability of their internal functioning. 
CoT reasoning has been also extended to multimodal LLMs 
\citep{hu2024visual, wu2024vstar, shao2024visual}: this presents new challenges compared to language-only domains, as the models need to attend to inputs from different modalities, and reason about both their individual content and how they are connected  \citep{awal2023investigating, kil2024iimmr, sun2025mitigating}.

Recent work has also begun to explore the integration of CoT into video understanding tasks, 
where the input to a multimodal LLM consists of a \emph{sequence} of images (the frames of the video) along with a text prompt. 
This makes reasoning on videos particularly complex, as the model needs to capture the semantics of the text prompt, understand temporal and causal relationships between frames, and reason about the video in its entirety.
Existing approaches rely on complex inference frameworks with architecture modifications \citep{hao2024video} or auxiliary networks \citep{han2024videoespresso} at evaluation time to extract and integrate key frames from the video sequence into the reasoning trace.
%into video LLMs by extracting key frames from the video sequence.
This makes using these models both more computationally expensive and less general (as they are specialized to some tasks), and deviates from the natural CoT prompting successfully applied to standard LLMs.
Another limitation of current methods is that collecting reasoning traces, needed for training, involves complex and costly procedures, which may limit how much the size of training datasets can be scaled in practice.
For example, \citet{wang2024videocot} iteratively refine reasoning traces via annotations from LLMs \emph{and} human experts, \citet{han2024videoespresso} leverage several auxiliary models, and \citet{hao2024video} use additional spatial-temporal scene graph data.
Finally, video LLMs do not currently provide explicit connections between particular segments of the video and their reasoning, which would better ground the chain-of-thought.

To remedy these limitations, in this work we propose \textbf{Chain-of-Frames (CoF)}, a new frame-aware chain-of-thought reasoning approach for video LLMs that 
integrates temporal information directly into a single-shot text-only CoT structure (see Fig.~\ref{fig:teaser}a). 
% ~\ref{fig:teaser-exp}).
% This enables the model to identify and refer to the most relevant frames while answering questions, in contrast to prior works that treat frame selection and reasoning as separate stages. 
Chain-of-frames is a simple and natural adaptation of the CoT paradigm in NLP to video understanding that does not require the auxiliary networks or complex inference frameworks of existing methods. Moreover, we propose an efficient data generation pipeline that allows us to collect a large dataset of CoF examples, named \cofdata.
To achieve this, a key element consists of leveraging a synthetic video dataset \citep{yi2020clevrer} to extract a large and diverse set of temporally grounded reasoning traces at virtually no cost.
Then, we fine-tune three recent open-source video LLMs, \ivls \citep{chen2024expanding}, \ivlb \citep{zhu2025internvl3} and Phi-3.5-Vision~\citep{abdin2024phi3} on our \cofdata.
In an extensive evaluation on five established benchmarks, we show that our CoF models significantly outperform the original ones with and without naive CoT prompting \citep{wei2022chain} (see Fig.~\ref{fig:teaser}b).
%-acc}).
Moreover, they are competitive with or better than state-of-the-art (SOTA) open- and closed-source models (Tab.~\ref{tab:summary_results}). Finally, further analysis reveals that synthetic data alone—despite being out-of-distribution relative to real-world benchmarks—provides a substantial improvement in model accuracy. This indicates that the model effectively learns the underlying task from the synthetic data and successfully transfers this knowledge to real-world scenarios.
% Notably, our CoF-\ivlb achieves higher accuracy than the best model reported on \videomme,
% \mvbench, and \vsi leaderboards, surpassing even proprietary models like GPT-4o and Gemini 1.5 Pro.
% These results also demonstrate that video LLMs can learn the ability to produce frame-aware reasoning from a limited number of samples and generalize the behavior to unseen tasks.
% Thus, our work provides a simple and inexpensive approach to adapt the CoT paradigm for video LLMs that improves both performance and interpretability. 

% \begin{figure*}[t]
%     \centering
%     \begin{minipage}[b]{0.63\linewidth}
%     \centering
%     \includegraphics[width=\linewidth, trim=50mm 0mm 0mm 0mm, clip]{figures/pl_cof_teaser.pdf} 
%     \subcaption{A CoF reasoning trace generated by our CoF-4B model.}
%     \label{fig:teaser-exp}
%     \end{minipage}
%     %
%     \hfill
%     %
%     \begin{minipage}[b]{0.35\linewidth}
%     \centering
%     \includegraphics[width=\linewidth, trim=0mm 0mm 0mm 0mm, clip]{figures/teaser_acc.pdf} 
%     \subcaption{CoF models vs the baseline models.}
%     \label{fig:teaser-acc}
%     \end{minipage}
%     \caption{\textbf{(a)} Chain-of-frames reasoning generated by our CoF-\ivls model: 
%     it includes the key frames 
%     to answer the question (from \videomme).
%     \textbf{(b)}
%     Comparison of accuracy across multiple video understanding benchmarks between baseline models (\ivls and \ivlb) and their CoF-enhanced counterparts: 
%     our models consistently outperform the baselines in nearly every case.
%     }
%     \label{fig:teaser}
% \end{figure*}

\begin{figure*}
    \centering
    \includegraphics[width=0.95\linewidth, trim=0mm 55mm 0mm 0mm, clip]{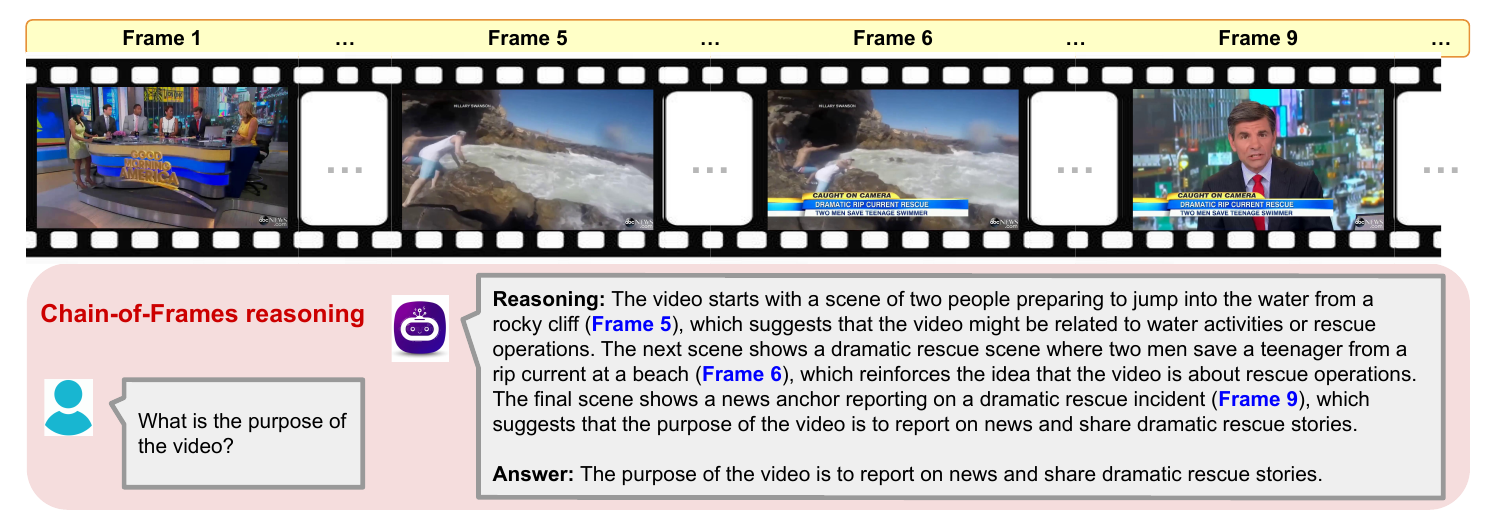}
    \includegraphics[width=\linewidth, trim=10mm 55mm 10mm 0mm, clip]{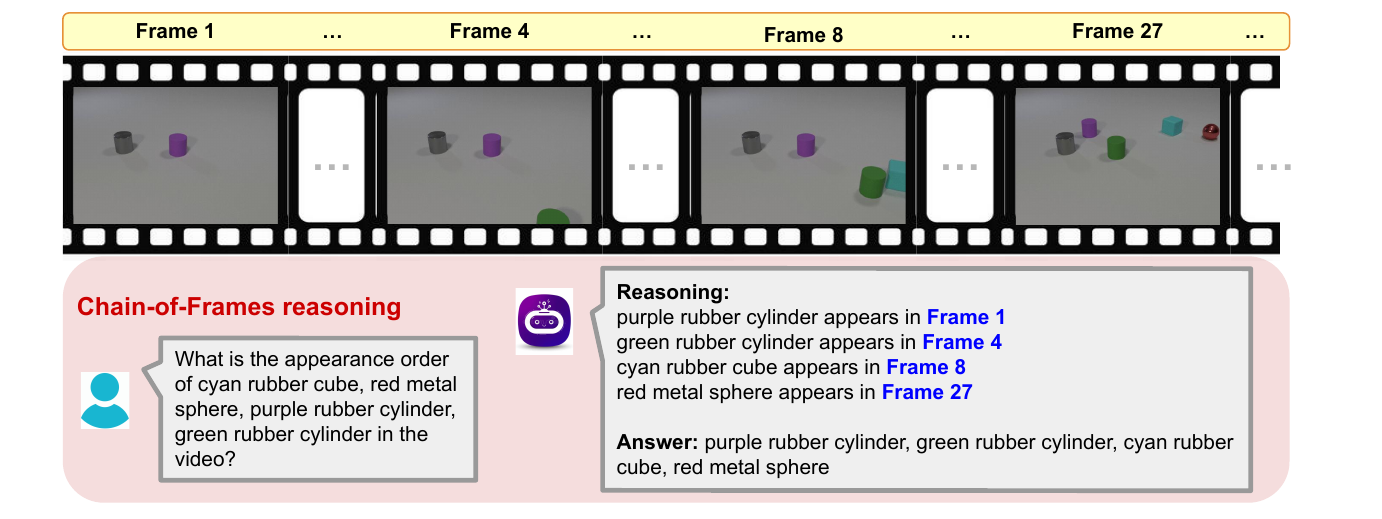}
    \vspace{-5mm}
    \caption{
    \textbf{\cofdata.} We show examples of our training data with chain-of-frames reasoning, including video, question, reasoning trace and answer. We include samples from \cofdatareal (real video, top row) and \cofdatasyn (synthetic video, bottom), created as described in Sec.~\ref{sec:data}.
    }
    \label{fig:train-data-example}
\end{figure*}

\section{Related Work}
\myparagraph{Multimodal LLMs for videos.}
Multimodal Large Language Models~\citep{chen2023internvl, li2024llava, ghazanfari2024emma} have made substantial progress in integrating visual and textual modalities, enabling them to perform complex reasoning and achieve deep understanding across various data types, including videos~\citep{chen2023internvl,xue2024longvila,li2023vila,zhang2024longva,zhang2024llavanextvideo,li2024llavaonevision,bai2024qwen2vl, zhu2025internvl3}. 
Among recent advancements, InternVL2.5~\citep{chen2024expanding}, InternVL3 \citep{zhu2025internvl3}, LLaVA-NeXT-Video~\citep{zhang2024llavanextvideo}, and Qwen2-VL~\citep{wang2024qwen2} stand out for their strong video understanding capabilities. Both InternVL and LLaVA-NeXT-Video process videos in an image-text interleaved format, 
aligning sequences of video frames with language to form a unified multimodal stream. In addition to these open-source models, closed-source systems such as GPT-4o~\citep{achiam2023gpt} and Gemini-1.5~\citep{team2024gemini} have also demonstrated impressive multimodal capabilities, although details of their architecture and training remain proprietary.

\myparagraph{Chain-of-Thoughts for videos.}
Recent research on chain-of-thought (CoT) reasoning for video understanding can be broadly categorized into two paradigms: (1) single-stage methods, which directly generate textual reasoning explanations followed by an answer in a single step~\citep{wang2024videocot}, and (2) multi-stage methods, which employ auxiliary models to extract meta-information in earlier stages and then use this information to generate the answer in the final step~\citep{han2024videoespresso,hu2025mllm,hu2025cos}.
In the first category, VideoCoT~\citep{wang2024videocot} introduces an active annotation framework to produce reasoning explanations from video descriptions, thereby guiding the model to generate reasoning traces followed by the answer.
In the second category, most approaches~\citep{wang2025videotree,hu2025mllm,hu2025cos, han2024videoespresso,buch2025flexible} emphasize key frame selection as an initial stage to enhance downstream answer generation. 
Building on this idea, VideoEspresso~\citep{han2024videoespresso} further integrates additional steps, combining the selected core frames with the question and feeding them into a reasoning model that performs chain-of-thought (CoT) reasoning, with the final answer produced in a subsequent stage.
Finally, Video-of-Thought~\citep{hao2024video} introduces a more complex five-stage pipeline that constructs spatio-temporal scene graphs to reason over videos and answer multiple-choice questions.
% M-LLM~\citep{hu2025mllm} leverages multimodal large language models to identify the most query-relevant frames, while Chain-of-Shot~\citep{hu2025cos} designs a prompting strategy tailored for long-form videos that selects key frames to support reasoning. 

\section{Chain-of-Frames: Reasoning on Videos via Frame References}
\subsection{Limitations of reasoning on videos}
Recent work has begun to explore the integration of 
CoT into video understanding tasks. While the base language models may be trained to produce reasoning traces, these are not specific for reasoning on videos.
To encourage chain-of-thought output in video LLMs, models must be fine-tuned on video-grounded reasoning traces. VideoCoT~\citep{wang2024videocot} addresses this by generating CoT training data through a pipeline that leverages video descriptions as raw annotations and uses LLMs to produce corresponding reasoning traces. However, because the descriptions are not temporally aligned at the frame level, the resulting reasoning traces can introduce temporal inconsistencies~\citep{li2025vidhalluc}. To mitigate such inaccuracies, human annotators are introduced into the loop to refine the LLM-generated reasoning traces.
As a result, the reliance on human and LLM annotations makes data generation expensive, limiting the dataset to only 11k samples.
Moreover, the traces lack explicit temporal grounding; that is, individual reasoning steps are not clearly aligned with the corresponding video frames.

A different strategy is to leverage multiple auxiliary models to generate the meta-information, including finding the key frames from the video and explicitly grounding the model's answer to the key frames~\citep{han2024videoespresso, hu2025mllm, hao2024video}. While these multi-stage pipelines improve video reasoning, their computational overhead during inference poses a major bottleneck. Furthermore, because only a subset of the frames is passed to the video LLM, the video’s full temporal context is lost.

In summary, current video reasoning methods face two major limitations: (1) they depend on expensive and often inaccurate training data generation while lacking explicit temporal grounding in the reasoning process, or (2) involve complex inference procedures that rely on auxiliary models and compromise contextual understanding by processing only a subset of the frames to generate the final answer.

\begin{figure*}
    \centering
\includegraphics[width=\linewidth, trim=0mm 10mm 0mm 0mm]{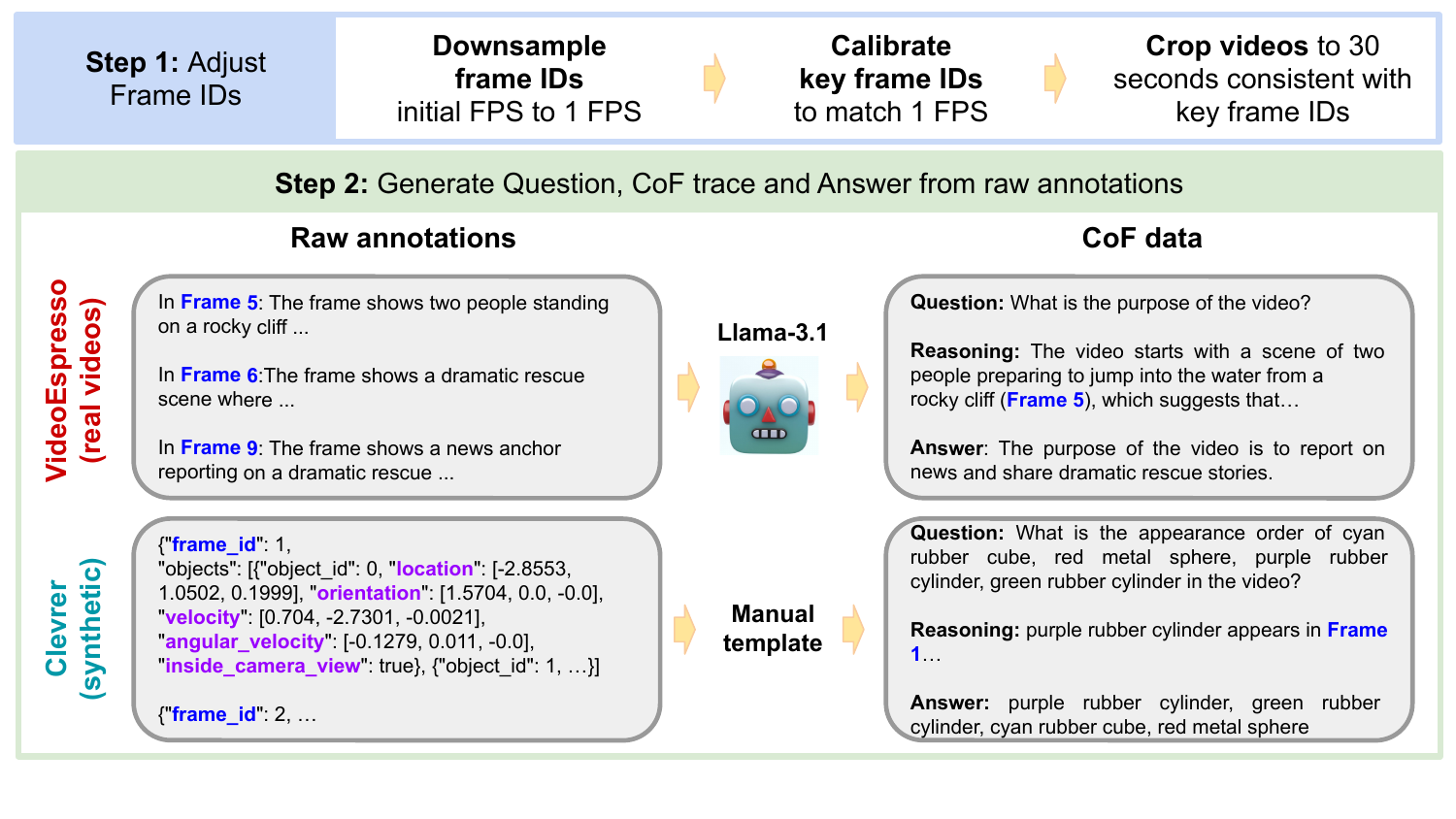}
\caption{\textbf{Overview of our two-step pipeline for generating \cofdata.} 
Step 1 adjusts the frame IDs while preserving frame-caption alignment.
Step 2 utilizes raw annotations to generate CoF triplets (question, frame-aware reasoning trace, answer), using Llama-3.1-8B-Instruct for real-video data and manual templates for synthetic data.}
\label{fig:data_generation}
\end{figure*}

\subsection{Chain-of-Frames}
To address these limitations, we propose \textbf{Chain-of-Frames (CoF)}, a simple yet effective approach introducing temporal grounding into the reasoning process. CoF consists of reasoning traces with explicit references to frames relevant to answering the given query, within a single-stage inference. 
Concretely, we use the position of the frame in the video  (e.g., ``Frame 1'', ``Frame 2'', ...) as an identifier. Unlike timestamps, this representation is agnostic to video duration and sample frequency, making it more consistent across diverse video data and potentially easier to learn. 
Examples are shown in Fig.~\ref{fig:teaser} and Fig.~\ref{fig:train-data-example}, with additional illustrations provided in App.~\ref{sec:additional_figures}.
This approach has several benefits:

\begin{enumerate}[left=0mm, itemsep=4pt, parsep=0pt, topsep=0pt]
    
    \item \textbf{Data quality:} as we show in Sec.~\ref{sec:data}, training data can be efficiently generated from existing annotated real videos, where the video descriptions are temporally aligned at the frame level, resulting in temporally accurate reasoning traces. Moreover, CoF traces can be generated from synthetic videos with high precision by combining their rich annotations with manually designed templates, at virtually no additional cost. This low-cost and scalable data generation process stands in contrast to the more complex pipelines~\citep{wang2024videocot, han2024videoespresso}.
    \item \textbf{Simplicity:} our reasoning traces are expressed entirely in natural language—without requiring auxiliary modules to generate complex data formats, bounding boxes, or scene graphs as in \citet{hao2024video, han2024videoespresso}. This makes our approach a natural extension of standard Chain-of-Thought (CoT) reasoning used in NLP tasks, eliminating the need for multi-stage inference pipelines and reducing overall latency.
    \item \textbf{Explicit temporal grounding:} temporal grounding is achieved by explicitly referencing the task-relevant frames within the reasoning traces, rather than by retrieving and processing key frames through a multi-stage inference process. This design strengthens the connection between visual content and reasoning, making CoF a temporally grounded extension of the typical CoT approach proposed by \citet{wang2024videocot}.
    \item \textbf{Interpretability:} %since CoF are in natural language and include references to specific frames, %in a similar way as humans would reason about videos, 
    The reasoning traces generated at inference time include frame references (Fig.~\ref{fig:teaser}), providing direct insights into how the LLM's decisions are obtained.
    %Additionally, this may reveal which
\end{enumerate}

% In the next section, we detail how we generate chain-of-frames from video datasets.
% These traces constitute the training data for fine-tuning video LLMs capable of producing CoF-based reasoning.

\begin{figure*}
    \centering
    \includegraphics[width=.9\linewidth]{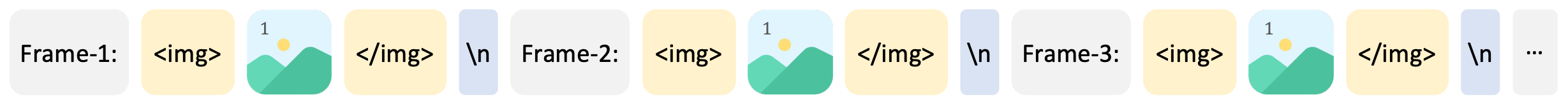}
    \caption{\textbf{Video encoding format.} InternVL models add a textual identifier before each frame, which is well-suited for our chain-of-frames reasoning. The illustration is taken from \citet{chen2024expanding}.}
    \label{fig:internvl-input}
    \vspace{-3mm}
\end{figure*}

\subsection{Chain-of-Frames training data collection %(\cofdata)
}
\label{sec:data}
We construct chain-of-frames traces from both real and synthetic videos. For real videos, we use the training split of the \videoespresso dataset \citep{han2024videoespresso}, which features videos from a wide range of sources and includes descriptions of key frames for each.
These diverse frame-level annotations provide a starting point for extracting CoF traces.
To complement this, we use the training split of the \clevrer dataset~\citep{yi2020clevrer}, which contains synthetic videos of simple 3D objects interacting within a controlled environment, along with rich annotations.
Synthetic data offers two key advantages: it overcomes the cost of generating training samples from real videos and provides highly accurate temporally grounded reasoning trace derived from the rich annotations with manual templates.
We next describe the main steps of our data generation pipeline, illustrated in Fig.~\ref{fig:data_generation}.
% , and introduce the resulting dataset of chain-of-frames, named \cofdata.
Additional details and examples are provided in App.~\ref{sec:experimental_details}.

\myparagraph{Frame ID alignment.} The original annotations include frame IDs, but, due to context length limitations of video LLMs, we downsample the videos while preserving the frame-annotation alignment. We first map each frame to its timestamp, and clip the video to the maximum duration allowed by the model (e.g., 30 seconds in our experiments), ensuring the segment includes all frames for which captions are available. We then re-calibrate the frame IDs to reflect their new positions within the clipped video.

\myparagraph{CoF from real videos (\cofdatareal).}
\videoespresso provides captions for key frames. After aligning frame IDs, we obtain data in the format shown in Fig.~\ref{fig:data_generation} (raw annotations).
From these temporally aligned annotations, we generate triplets of questions, answers, and reasoning traces with frame references by prompting an LLM using the raw annotations as input.
For this, we use Llama3.1-8B-Instruct~\citep{meta2024llama3.1} (the full prompt is provided in App.~\ref{sec:experimental_details}).
This process yields multiple questions per video, often covering diverse parts of the video and referencing different subsets of frames.
An example from \cofdatareal is in Fig.~\ref {fig:data_generation}, and a complete training sample including the video is in Fig.~\ref{fig:train-data-example}.

\myparagraph{CoF from synthetic videos (\cofdatasyn).}
In each frame of \clevrer, every object is annotated with both fixed properties (shape, material, color) and situational attributes (e.g., velocity, location), see Fig.~\ref{fig:data_generation}.
We use these rich attributes to generate three categories of quantitative questions (\textit{object count}, \textit{appearance order}, and \textit{relative distance}) which complement the semantic questions obtained for real video.
Notably, we can generate both questions, answers, and chain-of-frames using fixed manual templates (see App.~\ref{sec:experimental_details}), since all the necessary information can be directly deduced from the object-specific raw annotations.
This eliminates the need for using an LLM, significantly reducing generation cost and enabling easy scaling of the dataset size.
An example of an \textit{appearance order} question is shown in Fig.~\ref{fig:train-data-example} (%examples from 
for other categories %are shown in 
see App.~\ref{sec:additional_figures}). 
\\

\begin{figure}[H]
    \centering
\includegraphics[width=0.8\linewidth,
trim=10mm 15mm 10mm 0mm]{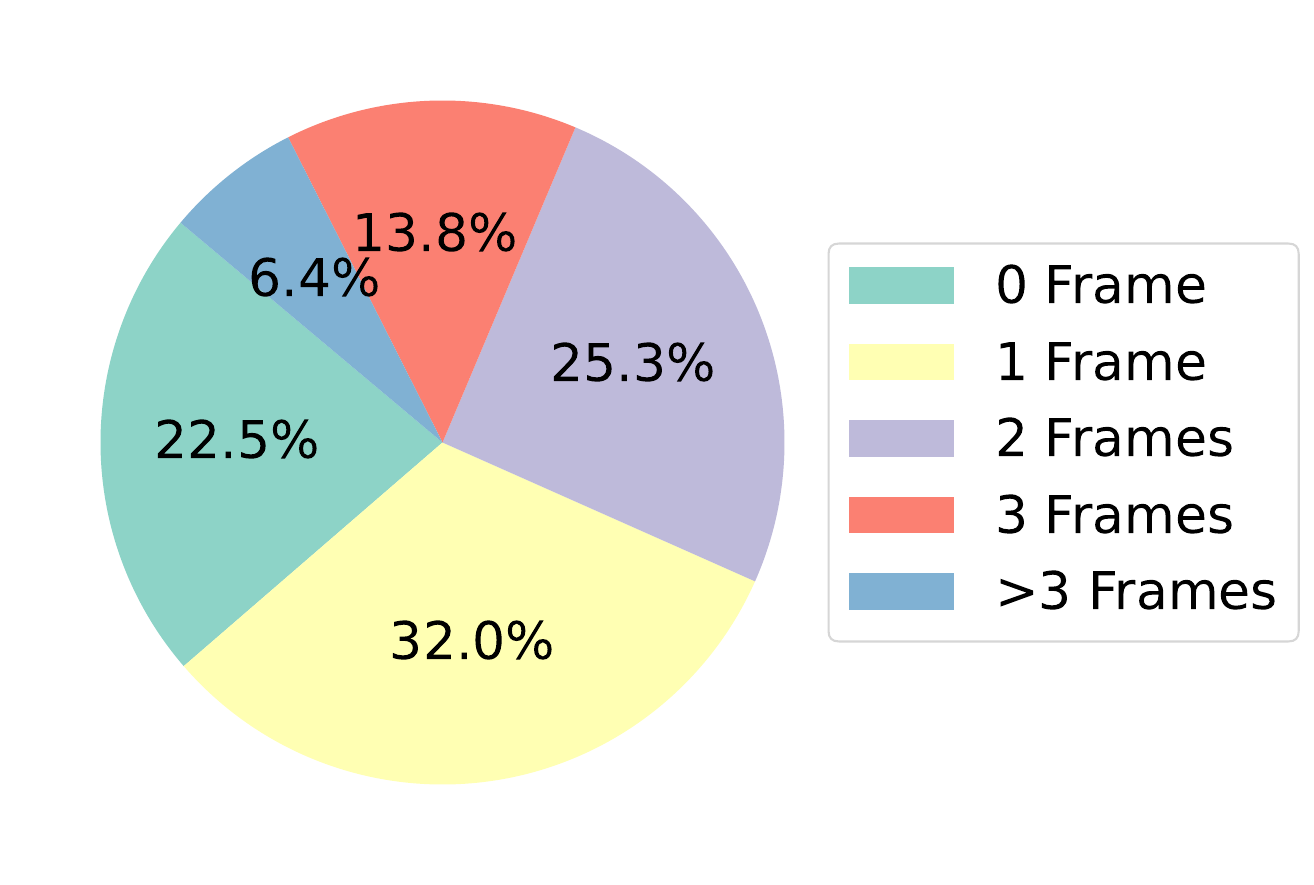}
    \caption{Distribution of frame references in \cofdata. %training data.
    } 
    \label{fig:train-data-stats}
\end{figure}

\myparagraph{Final dataset (\cofdata).} 
From the generated chain-of-frames, we filter out samples where frames are referred in the question (as this does not happen at test time).
Moreover, we reduce the number of samples with no frame references in the reasoning trace to give higher weight to more complex examples of reasoning.
We nevertheless keep a non-negligible fraction of samples with no frame references since there might be, in the evaluation benchmarks, questions which do not require CoF-like reasoning, and we do not want to force the model to generate it when unnecessary.
The resulting dataset comprises 164,186 samples. %, comprising 103,683 samples from the \cofdatareal dataset%, which is based on real-world videos,
%and 60,503 samples from the \cofdatasyn dataset of synthetic videos.
Fig.~\ref{fig:train-data-stats} illustrates the distribution of the number of frames referenced per sample.
\vspace{-4mm}
% Fig.~\ref{fig:train-data-stats} shows the distribution of how many frames are referenced in the reasoning traces for the final dataset.
% and the individual splits.
% The \cofdatasyn exhibits a more balanced distribution compared to the automatically generated \cofdatareal: 
% this highlights that using synthetic videos allows us to better control various aspects of the data.

\section{Experiments}
\label{sec:experiments}

\subsection{Experimental setup}

\begin{table*}[t]
    \centering
    \small
    \tabcolsep=1.5pt
    %\resizebox{\textwidth}{!}{%
    \begin{tabular}{L{33mm}|*{5}{C{20mm}} | C{20mm}}
        
     Model & \textbf{\vsi} & \textbf{\videomme} & \textbf{\mvbench} &  \textbf{\vidhal} & \textbf{\eventhallusionshort} & Average \\ %& \textbf{\nextqa} \\
     \toprule 
     \multicolumn{4}{l}{\textbf{Closed-source Models}} \\
     \midrule
     GPT-4V/4T  & - & 59.9 & 43.7 & - & 76.5 & -\\
     GPT-4o & 34.0 & 71.9 & - & 77.2 & \textbf{91.9} & - \\ % & 71.2\\
     Gemini-1.5-Pro & \underline{48.8} & \textbf{75.0} & - & 67.1 & \underline{80.4}& - \\% & 70.4\\
     \midrule
     \multicolumn{4}{l}{\textbf{Open-source Models}} \\
% 58.0
% 62.7
% 53.2
% 50.2
% 58.0
% 60.8
% 67.0

% 64.6
% 72.1
     \midrule
     LLaVA-OneVision-72B & 40.2 & 66.2 & 59.4 & 64.7 & 59.5 & 58.0 \\%& 80.9\\
     Qwen2-VL-72B & 37.6 & 71.2 & 73.6 & 76.2 & 54.7 & 62.7 \\% & 81.2\\
     LLaVA-OneVision-7B & 32.4 & 58.2 & 56.7 & 58.4 & 60.1 & 53.2 \\% & 79.4\\
     LLaVA-NeXT-Video-7B & 35.6 & 46.5 & 53.1 & 50.9 & 64.8 & 50.2 \\ % & 62.4\\
     Qwen2-VL-7B & 31.0 & 63.3 & 67.0 & 69.6 & 59.3 & 58.0 \\ %& 77.6\\
     \ivls & 33.5 & 54.7 & 71.5 & 77.0 & 67.4 & 60.8\\ %& 75.3\\
     \ivlb & 41.0 & 66.5 & 74.4 & \textbf{80.9} & 72.1 & \underline{67.0} \\ %& \underline{82.4}\\
     \midrule
     \multicolumn{4}{l}{\textbf{Our Models}} \\
     \midrule    
     CoF-\ivls & 36.9 & 59.7 & \underline{76.1} & 79.2 & 71.2 & 64.6 \\ %& 79.6\\
     CoF-\ivlb  & \textbf{51.3} & \underline{73.7} & \textbf{77.1} & \underline{79.5} & 78.7 & \textbf{72.1} \\%& \textbf{87.3}\\
     \bottomrule
    \end{tabular}%}
    \caption{\textbf{Comparison of CoF-models to state-of-the-art video LLMs.}
    We report accuracy on five benchmarks (and their average) for relevant baselines and our CoF-models fine-tuned on \cofdata.
    Both CoF-\ivls and CoF-\ivlb outperform the majority of baselines despite significantly fewer parameters, and obtain the best or second-best results on four out of five benchmarks.
    }
    \label{tab:summary_results}
\end{table*}

\myparagraph{Baseline models.} 
While chain-of-frames is a general approach, we find it particularly well-suited for the recent InternVL models \citep{chen2024expanding, zhu2025internvl3}.
These models introduce a novel format for videos, where frames are interleaved with text identifiers such as \texttt{Frame-1}, \texttt{Frame-2}, etc. (see Fig.~\ref{fig:internvl-input}), reinforcing the temporal structure of the video.
This format already associates the image with their corresponding textual identifiers, and using these identifiers in reasoning traces may facilitate long-range interaction in long-context LLMs.
The InternVL models achieve state-of-the-art performance among open-weight video LLMs, and thus improving their reasoning capabilities is both challenging and of practical relevance.
Consequently, we adopt the recently released \ivls~\citep{chen2024expanding} and \ivlb~\citep{zhu2025internvl3} models for our experiments.
To test the generalizability of our approach beyond the InternVL family, we use we employ Phi-3.5-Vision~\citep{abdin2024phi3}, a mobile-scale multimodal LLM that demonstrates strong performance in language reasoning, as a third baseline model.
For limited spaces, we present the results relative to Phi-3.5-Vision in App.~\ref{sec:additional_experiments}, where we show that CoF is also effective on this video LLM.
%Further details on the models training procedure and experimental results are provided in App.~\ref{sec:experimental_details} and App.~\ref{sec:additional_experiments}.
% We leave exploration of CoF on other architectures to future work.

\myparagraph{Training details.} For \ivls, we fully fine-tune both the LLM and projection modules, while keeping the vision encoder frozen. For \ivlb, we apply LoRA-based fine-tuning~\citep{hu2022lora} to reduce memory usage. 
We refer to the resulting fine-tuned models as CoF-\ivls and CoF-\ivlb throughout the paper.
Detailed training configurations can be found in App.~\ref{sec:experimental_details}.

\myparagraph{Inference details.} We uniformly sample 30 frames from each input video for inference. This approach ensures consistent temporal coverage regardless of video length. Importantly, the model is not restricted to 30-second videos and can process videos of arbitrary duration.

\myparagraph{Video benchmarks.}
We compare the video LLMs on five popular benchmarks that 
% capture diverse aspects of video understanding.
%
% These benchmarks 
span a broad range of tasks, video types and duration, providing a comprehensive evaluation of model capabilities.
\videomme~\citep{fu2024video} includes six visual domains, including videos from 2 to 60 minutes long. \mvbench~\citep{li2023mvbench} encompasses 20 tasks that require more than single-frame analysis.
\vsi~\citep{yang2024think} focuses on quantitative reasoning tasks like object counting or appearance order.
To evaluate hallucination in video LLMs, we %additionally 
include \vidhal~\citep{choong2024vidhal} and \eventhallusion~\citep{zhang2024eventhallusion}.
Details on each benchmark and the breakdown of the results over the fine-grained splits are available in App.~\ref{sec:experimental_details} and App.~\ref{sec:additional_experiments}.

% Between the two individual datasets, the model trained on synthetic videos outperforms the one trained on real videos in three out of five cases, while being worse in just one.
% This suggests that further improvements might come from expanding the tasks covered by the synthetic data, which can even be generated at low cost.

\subsection{Main experiments}

We now contextualize the performance of our CoF-based models in relation to the current leading video LLMs.
For this, in Tab.~\ref{tab:summary_results} we report the performance of some of the strongest both closed-source (GPT-4V/4T~\citep{achiam2023gpt}, GPT-4o~\citep{hurst2024gpt}, Gemini-1.5-Pro~\citep{team2024gemini}) and open-source
(Qwen2-VL-72B~\citep{wang2024qwen2}, LLaVA-OneVision-72B~\citep{li2024llavaonevision}) models of different sizes.
We could not add the prior video CoT models~\citep{wang2024videocot, han2024videoespresso, hao2024video, hu2025mllm} in Tab.~\ref{tab:summary_results} as these are not publicly available.
Nevertheless, Tab.~\ref{tab:additional_baseline} shows comparisons on shared benchmark datasets, where our CoF-based models outperform these baselines. 
Finally, we note that we do not refer to the results of any models on our benchmarks as zero-shot. In fact, for closed-source and some open-source models, the training data is not known. Moreover, 5 out 20 tasks of \mvbench are partially based on the test set of \clevrer and \vidhal includes some videos from \mvbench.
% \vspace{-4mm}

\myparagraph{Comparison to base models.} 
Tab.~\ref{tab:summary_results} shows that CoF-enhanced models consistently outperform their corresponding base models across nearly all benchmarks.
CoF-\ivlb achieves 5.1\% higher average performance than the original \ivlb, while the CoF-\ivls improves of 3.8\% over \ivls. %(even when CoT prompting is used for the baselines)
Moreover, the fact that performance gains are more notable for the larger model suggests that more capable LLMs may benefit more substantially from CoF-based fine-tuning.

\myparagraph{Comparison to state-of-the-art video LLMs.} 
% , and, e.g., InternVL2.5 models have been pre-trained on \clevrer.
Tab.~\ref{tab:summary_results} also illustrates that, despite being significantly smaller than most baselines, CoF-\ivlb achieves the best results on \vsi and \mvbench, outperforming leading (closed-source) video LLMs.
It also has second-best accuracy on  \videomme and \vidhal, and best among open-source LLMs on \eventhallusion.
Meanwhile, CoF-\ivls attains competitive results, surpassing several larger open-source models and in some cases even closed-source ones.
These results demonstrate the effectiveness of chain-of-frames reasoning in improving video understanding across diverse and challenging benchmarks.
%
% Finally, in App.~\ref{sec:additional_experiments} we report the breakdown of the results over the subsets of the benchmarks in Table~\ref{tab:summary_results}.
% From those we can see, for example, that training with CoF leads to improvements in the spatial reasoning tasks of \vsi.
% We hypothesize that this is due to having the reasoning traces about object count, appearance order, and relative distance created from \clevrer in \cofdata.
% This suggests that the model is effectively learning spatial reasoning tasks from synthetic CoF data.

\myparagraph{Comparison to video CoT methods.}
Prior video CoT models~\citep{wang2024videocot, han2024videoespresso, hao2024video, hu2025mllm} are not accessible to the public, and they do not report results on the five benchmarks we use.
However, we provide comparative analyses to M-LLM~\citep{hu2025mllm} and Video-of-Thought~\citep{hao2024video} on \videomme and \nextqa~\citep{xiao2021next} using the results they report. Both baselines adopt multi-stage inference that explicitly retrieves key frames, whereas our approach integrates frame retrieval implicitly through frame references within the reasoning traces. In Tab.~\ref{tab:additional_baseline}, we see that our CoF-based models, including CoF-\ivls, which has fewer parameters than the competitors, outperform these baselines on both benchmarks.
Moreover, compared to the most recent model, M-LLM, our method achieves a greater improvement (4.9\% vs 0.8\%) on \nextqa despite starting from a stronger base model.

%\section{Analyses}
\section{Additional analyses} % of Chain-of-Frames}

\subsection{Influence of synthetic training data}

\begin{figure*}[t]
    \centering
    \includegraphics[width=.8\linewidth]
    {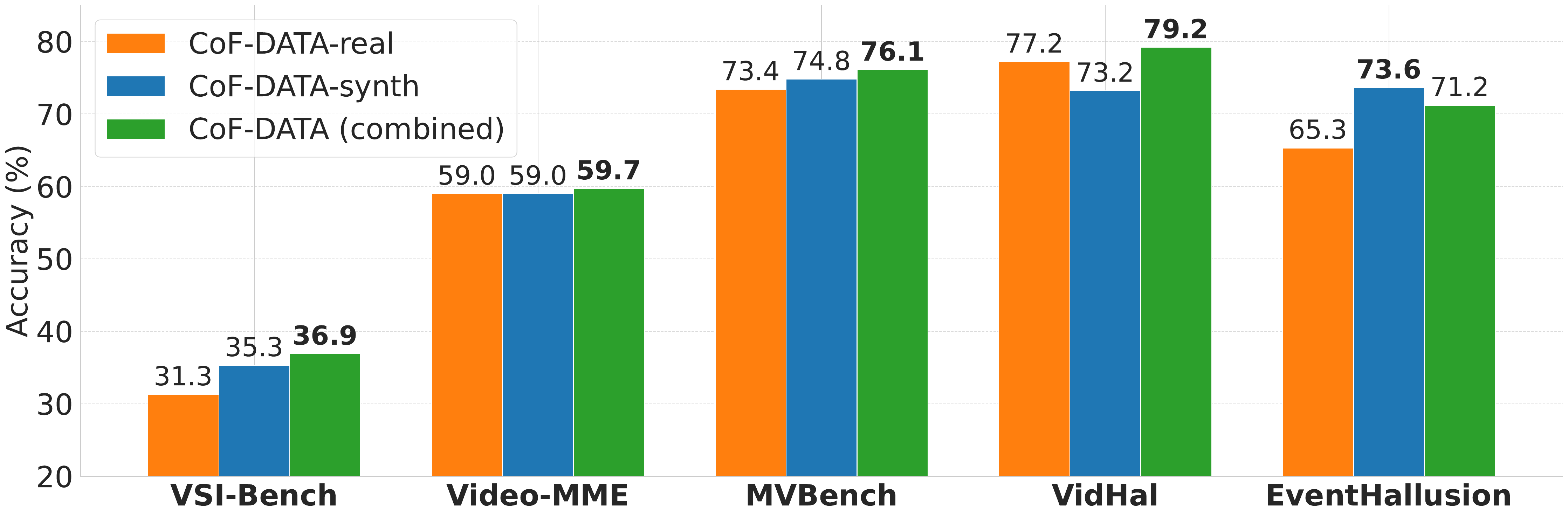}
    \caption{\textbf{Influence of synthetic training data.} Across most benchmarks, the model trained on synthetic data consistently outperforms the one trained on real data. Despite the substantial distribution gap between the synthetic samples and the real frames in these benchmarks, the model effectively learns the tasks from synthetic data and transfers this knowledge to real-world scenarios.
    For all models we use 164k training samples, and fine-tune an \ivls model.}
    % \caption{\textbf{Diversity of training data.} Combining CoF examples extracted from real and synthetic videos provides better results on almost all benchmarks compared to using a single source. For all models we use 164k training samples, and fine-tune an \ivls model.}
    \label{fig:data_abl}
\end{figure*}

The training data of \cofdata combines chain-of-frames from real and synthetic videos. As mentioned in Sec.~\ref{sec:data}, the synthetic portion of the dataset, \cofdatasyn, is generated without relying on heavy models, but rather through human-designed templates. The key question is how effective this synthetic data is for model training. 

%To evaluate the impact of dataset diversity, 
To address this question, we construct datasets of equal size, each sourced exclusively from either \cofdatareal or \cofdatasyn, and fine-tune \ivls on them. We then compare the resulting CoF-\ivls models in Fig.~\ref{fig:data_abl}.
First, we observe that across most benchmarks, the model trained on synthetic data consistently outperforms the one trained on real data. This result is notable: despite the considerable distributional gap between the synthetic samples and real frames from the benchmarks, the model effectively learns tasks such as \textit{object counting} and \textit{appearance ordering} from synthetic examples. Since synthetic data can be generated at low cost and large scale, this finding highlights a promising and scalable strategy for improving performance. Second, using the combined dataset outperforms single-source datasets on all benchmarks except for \eventhallusion, demonstrating the importance of diversity in the reasoning traces used for training.

\subsection{Chain-of-frames vs chain-of-thoughts variants}

\label{sec:reasoning_variants}
\begin{table*}[t]
\begin{minipage}{0.48\textwidth}
    % \begin{table}[t]
    \centering
    \tabcolsep=5pt
    % \vspace{2mm}
    \resizebox{\linewidth}{!}{%
    \begin{tabular}{ll|ll}
    %{L{26mm} L{26mm}|*{1}{L{20mm}} L{16mm}}
    Backbone & Model & \textbf{\videomme} & \textbf{\nextqa}  \\
     \toprule 
     %\multicolumn{3}{l}{\textbf{Video-LLaVA-7B}} \\
     %\midrule
     \multirow{2}{*}{Video-LLaVA-7B}
     & Original & 39.9 & 66.3 \\
     & Video-of-Thought \citep{hao2024video}  & - & 76.0\; \upblue{9.7}  \\
     \midrule
     %\multicolumn{3}{l}{\textbf{Qwen2-VL-7B}} \\
     %\midrule
     \multirow{2}{*}{Qwen2-VL-7B} & Original  & 58.1 & 77.6 \\
     & M-LLM \citep{hu2025mllm} & 58.7\; \upblue{0.6} & 78.4\; \upblue{0.8} \\
     \midrule
    %\multicolumn{3}{l}{\textbf{\ivls}} \\
     %\midrule
     \multirow{2}{*}{\ivls} & 
      \multirow{1}{*}{Original} & 54.9 & 75.3\\
     & SFT with CoF (ours) & 59.7\; \upblue{4.8} & 79.6\; \upblue{4.3} \\
    \midrule
    %\multicolumn{3}{l}{\textbf{\ivlb}}\\
    %\midrule    
    \multirow{2}{*}{\ivlb} & \multirow{1}{*}{Original} 
     & 66.5 & 82.4 \\
    % \cmidrule{2-3}
     & SFT with CoF (ours) & \textbf{73.7}\; \upblue{7.8} & \textbf{87.3}\; \upblue{4.9}\\
     \bottomrule
     
    \end{tabular}}
    \caption{\textbf{Comparison to video CoT methods.} Since the models from \citet{hu2025mllm} and \citet{hao2024video} are not publicly available, we compare the results they report on \videomme and \nextqa to those of our CoF models. 
    We also report the improvement of each approach over their baseline. Both our CoF models outperform the existing approaches on the two benchmarks.}
    \label{tab:additional_baseline}
% \end{table}

\end{minipage}\hfill
\begin{minipage}{0.48\textwidth}

    \centering
    \tabcolsep=1.1pt
    \resizebox{\linewidth}{!}{%
    \begin{tabular}{L{34mm}|*{6}{C{12mm}}}
        
     Model &  \rotatebox{45}{\textbf{\vsi}} & \rotatebox{45}{\textbf{\videomme}} & \rotatebox{45}{\textbf{\mvbench}} &  \rotatebox{45}{\textbf{\vidhal}} & \rotatebox{45}{\textbf{\eventhallusionshort}} \\
     \toprule 
     Original & 31.8 & 54.9 & 70.8 & 74.0 & 62.5\\
     + CoT Prompting & 33.5 & 54.7 & 71.5 & 77.0 & 67.4\\
    + SFT with QA only & 31.8 & 54.5 & 73.4 & 64.1 & 57.7\\
    + SFT with CoT & 34.3 & 58.6 & 73.7 & 77.9 & 53.1\\
    + SFT with CoF (ours) & \textbf{36.9} & \textbf{59.7} & \textbf{76.1} & \textbf{79.2} & \textbf{71.2}\\
         \bottomrule
    \end{tabular}}
    \caption{\textbf{Chain-of-Frames vs other Chain-of-Thoughts variants.}
    We compare different approaches to encourage reasoning in video LLMs, via either prompting or supervised fine-tuning (SFT), see Sec.~\ref{sec:reasoning_variants} for details. All models are obtained from \ivls. Fine-tuning on our chain-of-frames (CoF) data yields the best accuracy on all benchmarks.
    }
    \label{tab:reasoning_methods}
\end{minipage}
\end{table*}

\myparagraph{Setup.} 
We aim to evaluate how our chain-of-frames approach compares to alternative methods for incorporating reasoning into video LLMs, including both prompting and fine-tuning baselines.
To ensure consistency, we use \ivls as a baseline model, and for variants requiring fine-tuning, we adopt the training scheme detailed above.
We compare the following models.
\begin{itemize}[left=0mm, itemsep=0pt, parsep=4pt, topsep=4pt]
    \item \textbf{Original}: \ivls with default prompting.
    
    \item \textbf{Original + CoT Prompting}: the \ivls model with a prompt that encourages the model to perform intermediate reasoning before answering the question (see prompt in App.~\ref{sec:experimental_details}).
    This approach is similar to the standard CoT used in language tasks~\citep{wei2022chain}.
    
    \item \textbf{Original + SFT with QA only}:  the  \ivls model fine-tuned on the question-answers pairs from our \cofdata without including the reasoning traces. %supervised fine-tuning (SFT) of \ivls on the question-answers pairs from our \cofdata but excluding the reasoning traces.
    
    \item \textbf{Original + SFT with CoT}: the  \ivls model fine-tuned on \cofdata,
    where reasoning traces are included but references to specific frames are removed (e.g., \textit{``In Frame 1...''}  is replaced with a generic \textit{``In the video...''}). 
    % This approach mimics the standard CoT format. 
    % and is similar to the method of \citet{hao2024video}.
    % As their model is not publicly available, this method provides a direct comparison to their approach (VideoCoT) on our more comprehensive dataset.

    \item \textbf{Original + SFT with CoF}: the \ivls model fine-tuned on \cofdata, i.e. our proposed approach.
\end{itemize}
For all baseline models based on supervised fine-tuning (SFT), we report results using the best prompting strategy (either standard or CoT) for each benchmark. 
For our SFT with CoF model, we always use CoT prompting across all benchmarks. A complete comparison is provided in App.~\ref{sec:additional_experiments}. 

\myparagraph{Results.}
We report results for all models on the five benchmarks in Tab.~\ref{tab:reasoning_methods}.
First, we observe that CoT prompting alone already improves the accuracy of the original model 
% on four out of five benchmarks 
compared to standard prompting. 
This result highlights the value of encouraging explicit reasoning, which video LLMs are capable of performing.
Second, fine-tuning on question-answer pairs without reasoning (SFT with QA only) gives mix results, possibly due to overfitting on the training data, which might degraded the reasoning ability of the original model.
Training on reasoning traces without temporal grounding (SFT with CoT) improves the results across nearly all benchmarks, with the notable exception of \eventhallusion.
Finally, our proposed model, CoF-\ivls, fine-tuned on the CoF reasoning traces, achieves the highest accuracy across all benchmarks, with improvements ranging from 4.8\% to 8.7\% over the original \ivls.
These results demonstrate that our explicit temporal ground approach, chain-of-frames, offers consistent benefits across diverse tasks and domains and can serve as a strong alternative to key-frame-retrieval methods.

\subsection{CoF reasoning at inference time}
\begin{figure}[H]
    \centering
    \includegraphics[width=0.7\linewidth]{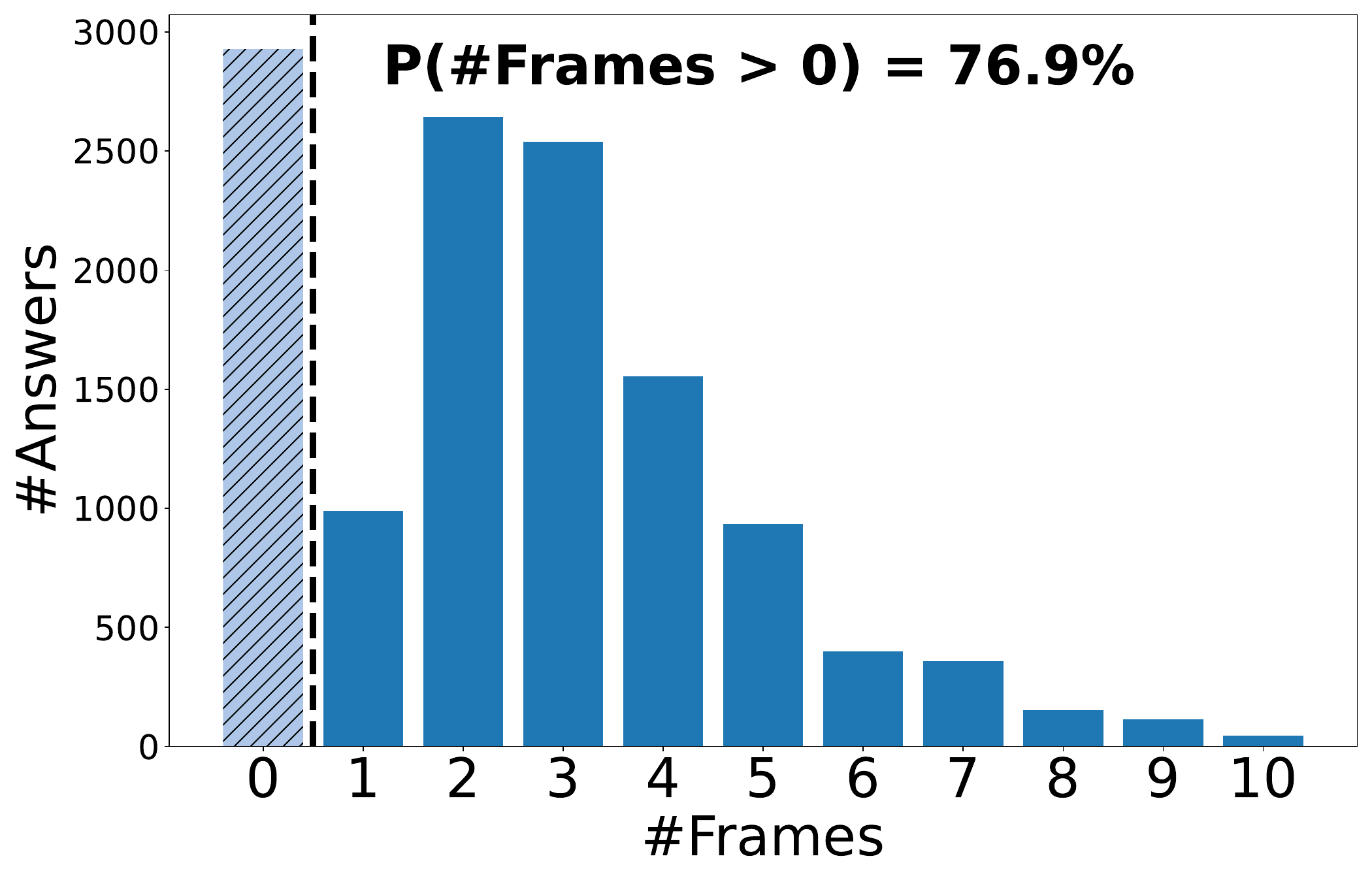}
    \caption{\textbf{CoF reasoning at inference time.} We show the distribution of the number of frame references generated by CoF-\ivlb at inference time.}
    \label{fig:frame-analysis}
\end{figure}

To better understand how the CoF models utilize frame references during inference, we track the number of frames referenced in the reasoning trace per answer.
In Fig.~\ref{fig:frame-analysis}, we report how many answers, generated by our CoF-\ivlb model across all evaluation benchmarks, contain $N$ frames for $N=1, \ldots, 10$.
We can see that our model learns to produce, from a relatively small training set, reasoning traces which refer to frame IDs (see qualitative examples in App.~\ref{sec:additional_figures}).
% In particular, CoF-\ivls includes at least one frame reference in 68.7\% of the answers, while CoF-\ivlb does it in 76.9\% of the cases.
% This suggests that larger models might benefit more from fine-tuning with frame-grounded reasoning: in fact, the CoF-\ivlb model shows both higher use of frame references and larger gains in accuracy (see Fig.~\ref{fig:teaser-acc}).
Moreover, the model can use frame references selectively rather than uniformly across tasks (distribution by benchmark in App.~\ref{sec:additional_experiments}), depending on the type of questions.

\vspace{-4mm}
\section{Conclusion}

We have introduced chain-of-frames (CoF), a new approach to encourage video LLMs to produce temporally grounded reasoning before providing the final answer.
Compared to existing works, CoF does not require complex ad-hoc inference frameworks or auxiliary models, and we show that its training data can be extracted more efficiently and accurately from both real and synthetic videos.
Our models fine-tuned on \cofdata outperform across multiple benchmarks those obtained with alternative methods for reasoning, and even achieve results better than or similar to leading video LLMs.
Overall, these features make CoF a viable option to further improve the reasoning capabilities of video LLMs.
Exploring the effect of increasing the size and diversity of the training data, as well as the scale of the models, represents an exciting direction for future work. 

\myparagraph{Limitations.} 
Our CoF training data is suitable for finetuning video LLMs that operate with a fixed FPS, ensuring that the frame indices referenced in reasoning traces align with the video frames provided to the language model. Consequently, an open question is how to adapt our method to video LLMs—such as Qwen2.5-VL~\citep{bai2024qwen2vl}—that employ dynamic frame rates, while preprocessing the video.
% We have currently applied CoF on the InternVL models since their video encoding is well-suited for our reasoning with frame references, and they achieve state-of-the-art performance.
% It is an interesting open question how to customize our approach to other types of video LLMs.
\section*{Acknowledgements}  
This paper is supported in part by the Army Research Office under grant number W911NF-21-1-0155 and by the New York University Abu Dhabi (NYUAD) Center for Artificial Intelligence and Robotics, funded by Tamkeen under the NYUAD Research Institute Award CG010. Additional support was provided by the NYU IT High Performance Computing resources, services, and staff expertise.
Moreover, F.C. and N.F. acknowledge support from an unrestricted gift from Google and by the Swiss National Science Foundation (grant number 212111).

{
\small
\bibliographystyle{ieeenat_fullname}
\bibliography{literatur}
}

\newpage
\clearpage
\appendix

\section{Experimental Details}
\label{sec:experimental_details}

\subsection{Chain-of-Frames training data}
\label{app:cof-data}

\myparagraph{CoF from real videos (\cofdatareal).}
To generate question-reasoning-answer triplets, we prompt Llama-3.1-8B-Instruct~\citep{meta2024llama3.1} using the instruction shown in Fig.~\ref{fig:prompt_llama} along with frame-aware video captions from the \videoespresso dataset (see Fig.~\ref{fig:data_generation} for details). Notably, the raw video content is not included in this process. Two examples from \cofdatareal are shown in Fig.~\ref{fig:cof_videoespresso}.

\paragraph{CoF from synthetic videos (\cofdatasyn).} The second portion of our training dataset is derived from the \clevrer dataset, which includes detailed attributes for each object in every video frame.
Specifically, given a frame ID and object ID, the \texttt{inside\_camera\_key} field indicates whether the object is visible in the frame, enabling us to determine when an object enters or exits the scene. The \texttt{velocity} attribute reflects whether an object is moving or stationary, while the \texttt{location} attribute provides its absolute or relative position, which can be leveraged to estimate distances or identify collisions.
The final \cofdatasyn dataset comprises three categories of questions: \textit{Object Count}, \textit{appearance order}, and \textit{relative distance}. 
Within the \textit{object count} category, we define three subtypes: \textit{(i)} \textit{collision-based} (“How many collisions...”), \textit{(ii)} \textit{motion state} (“How many moving objects...”), and \textit{(iii)} \textit{temporal-based}, where questions reference specific segments of the video (“After object A enters...”).
The questions, answers and reasoning traces are generated with the manually written templates shown in Fig.~\ref{fig:templates_clevrer}, making the data collection process particularly simple and fast.
Examples from \cofdatasyn are shown in Fig.~\ref{fig:cof_clevrer}.
\begin{figure*}[t]
    \centering
\includegraphics[width=0.8\linewidth]{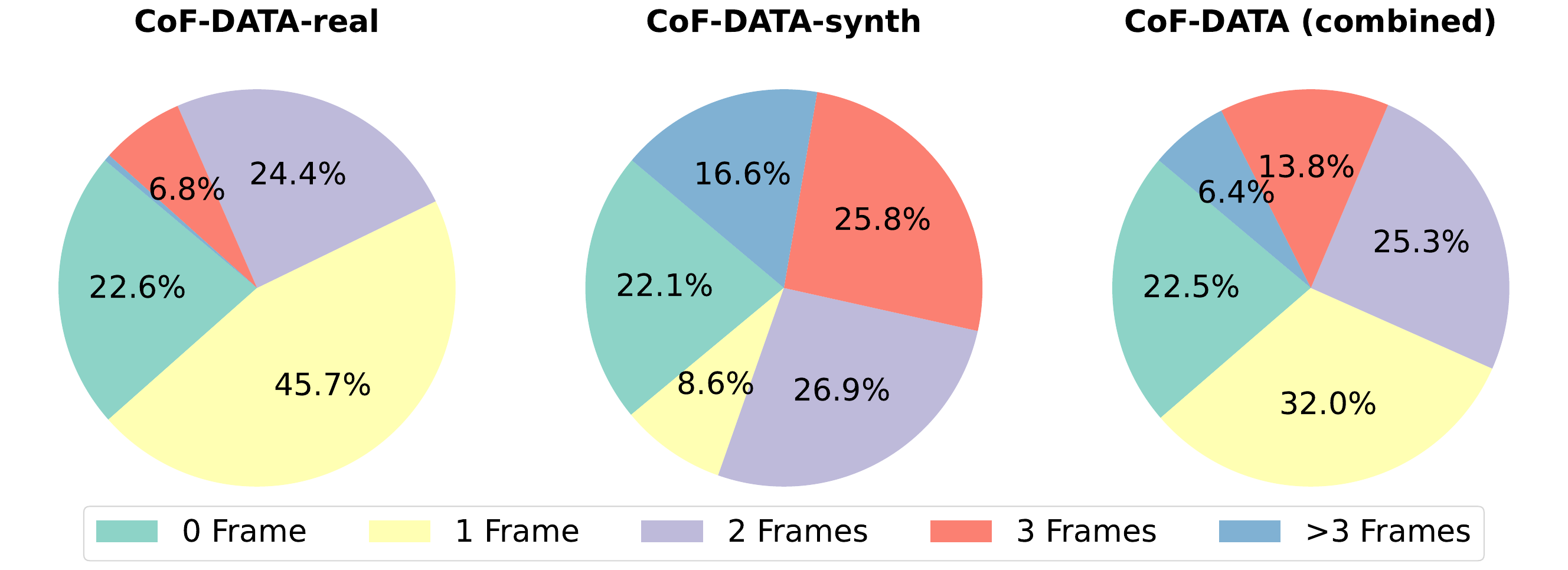}
    \caption{\textbf{Distribution of frame references in the Chain-of-Frames training data}. The left pie chart illustrates the distribution for \cofdatareal, having fewer frames per reasoning trace, whereas CoF-DATA-synth demonstrates a more balanced frame distribution due to controlled synthetic video generation. The right pie chart shows the overall distribution for the \cofdata.} 
    \label{fig:train-data-stats-full}
\end{figure*}

\paragraph{Final dataset (\cofdata).} Our training data, \cofdata, has a total of 164,186 samples, comprising 103,683 samples from the \cofdatareal dataset, which
is based on real-world videos, and 60,503 samples from the \cofdatasyn dataset of synthetic videos. Fig.~\ref{fig:train-data-stats-full} shows the distribution of how many frames are referenced in the reasoning traces, both
for the final dataset and the individual splits. The \cofdatasyn exhibits a more balanced distribution compared to the automatically generated COF-\cofdatareal: this highlights that using synthetic videos allows us to better control various aspects of the data.

\begin{figure*}[p]
\begin{promptbox}[Prompt]
Ask a question based on the narrative that is provided for a video. The questions should be answerable from the video description. 
Start reasoning step-by-step like this:
Point out key elements from the video relevant to the question. 
Break down the reasoning from those elements to the answer.
Include specific frame numbers as references to support your reasoning. 
Answer clearly. 
**Question**: 
**Reasoning**: 
**Answer**:
\end{promptbox}
\vspace{-2mm}
\caption{\textbf{Prompt for \cofdatareal.} We prompt Llama-3.1-8B to generate questions, answers, and reasoning traces with reference frames from the real videos of \videoespresso. Notably, to generate our training data, we do not use the videos but only their captions.}
\label{fig:prompt_llama}
\end{figure*}

\begin{figure*}[p]
    \centering
\begin{promptbox}[\textit{Object Count} Template]
Question: How many collisions happen in this video?

Reasoning: 
1. A collision happens in Frame <frame_id1> between <obj1_name> and <obj2_name>
2. ...
Answer:<#collisions> collisions happen in this video.
\end{promptbox}

\begin{promptbox}[\textit{Appearance Order} Template]
Question: what is the appearance order of <object_list> in the video?

Reasoning: 
1. <obj1_name> appears in Frame {frame_id}
2. ...
Answer: <sorted_object_list>
\end{promptbox}

\begin{promptbox}[\textit{Relative Distance} Template]
Question: Measuring from the closest point of each object, when <obj_name_t> <action> the scene, 
which of these objects (<all_objects_in_the_scene>) is closest to the <obj_name_t>?

Reasoning:  
1. <obj_name> <action> the scene in Frame <frame_id>. In Frame <frame_id>, the distance between <obj_name_t> and <obj_name_i> is <distances[t][i]>.
2. ...
Answer: <obj_name_{min(distances[t])}> 
% is the closet object to <obj_name_t> 
\end{promptbox}
\vspace{-2mm}
\caption{\textbf{Templates for \cofdatasyn.} To generate questions, answers and reasoning traces with reference frames from the annotations of the synthetic videos of \clevrer we rely on fixed, manually written templates. We create three types of questions (object count, appearance order, relative distance) with different templates.}
    \label{fig:templates_clevrer}
\end{figure*}

\subsection{Video benchmarks}
\label{app:exp-bench}

\paragraph{\videomme.} \videomme \citep{fu2024video} offers a diverse range of video types, covering six primary visual domains and 30 subfields to support broad scenario generalizability. It also introduces variation in temporal length, including short (under 2 minutes), medium (4-15 minutes), and long (30-60 minutes) videos.

\paragraph{\mvbench.} \citet{li2023mvbench} presents a comprehensive benchmark for multimodal video understanding, encompassing 20 challenging tasks that require more than single-frame analysis. It is specifically designed to evaluate a model's ability to understand temporal dynamics across video sequences.

\paragraph{\vsi.} This benchmark \citep{yang2024think} is designed to quantitatively assess the visual-spatial intelligence of multimodal large language models. Built from over 5,000 high-quality question-answer pairs across 288 real-world indoor videos, \vsi spans diverse environments such as homes, offices, and industrial spaces.
The benchmark covers eight tasks: object count, relative distance, relative direction, route planning, object size estimation, room size estimation, absolute distance estimation, and appearance order.
Out of these tasks included in this benchmark, relative distance, appearance order, relative directory, and route planning come with multiple-choice questions while the other four require an open-ended quantitative answer.
To better evaluate the proximity of the model's prediction with the correct answer, \cite{yang2024think} proposes using mean relative accuracy ($\mathcal{MRA}$).
Given a model's prediction $\hat{y}$ and ground truth $y$, relative accuracy is calculated by:
\begin{equation*}
    \mathcal{MRA} = \frac{1}{10} \sum_{\theta \in \mathcal{C}} \mathds{1} \left( \frac{|\hat{y} - y|}{y} < 1 - \theta \right)
\end{equation*}
where $\mathcal{C}=\{ 0.5, 0.55, \cdots, 0.95\}$ and denotes a range of confidence thresholds $\theta$ to calculate the relative accuracy.

\paragraph{\vidhal.} To evaluate video-based hallucinations in video LLMs, we use \vidhal \citep{choong2024vidhal}, a multiple-choice benchmark that features video instances drawn from public video understanding datasets, covering a diverse array of temporal concepts and aspects such as entity actions and event sequences. 

\paragraph{\eventhallusion.} \citet{zhang2024eventhallusion} introduce \eventhallusion, from which we use the binary-choice questions designed to systematically assess event-related hallucinations in video LLMs. From a hallucination attribution standpoint, it is specifically curated to evaluate a model’s susceptibility to language priors and vision-language correlation biases.

\begin{figure*}[t]
    \centering
\begin{promptbox}[CoT Prompting]
Given a video and a question, Start reasoning step-by-step like this:
Point out key frames from the video relevant to the question.
Break down the reasoning from those frames to the answer.
Conclude your reasoning to the answer.

Question: <question>
\end{promptbox}
\vspace{-2mm}
\caption{\textbf{CoT prompt.} We show the prompt used for elicit reasoning for both the baseline and our fine-tuned models.}
    \label{fig:cot-prompt}
\end{figure*}

\subsection{Chain-of-Frames model}
\label{app:cof-model}

\paragraph{CoF-InternVL.}
For \ivls, we fully fine-tune both the LLM and the projection modules, keeping the vision encoder frozen. In contrast, for \ivlb, we adopt LoRA-based fine-tuning~\citep{hu2022lora} to reduce memory consumption. All other training configurations remain consistent across both models. Training is conducted on a single H100 node equipped with 4 GPUs, using a learning rate of $2 \times 10^{-6}$, a batch size of 2, and a single epoch.

\paragraph{CoF-Phi-3.5-Vision-4B.}
To test the generalizability of our approach beyond the InternVL family, we employ Phi-3.5-Vision-4B~\citep{abdin2024phi3}, a mobile-scale multimodal LLM that demonstrates strong performance in language reasoning, as a third baseline model.

Phi-3.5-Vision-4B uses distinct, indexed image placeholder tokens such as <image$-i$>, instead of repeating a generic <image> token, because the model must uniquely align each image embedding with a specific position in the text sequence. Each placeholder is a separate token in the tokenizer, allowing the model to reliably map image embeddings to their corresponding locations and to correctly understand references to the images.

Phi-3.5-Vision's capability to understand references to the images makes it a good candidate for our base model. Similar to \ivls, we fully fine-tune both the LLM and the projection modules, keeping the vision encoder frozen. The other training configurations remain consistent across all baselines.

\section{Additional Experiments}
\label{sec:additional_experiments}

\paragraph{CoF-Phi-3.5-Vision-4B results.} 
Tab.~\ref{tab:prompt-analysis} compares the baseline Phi-3.5-Vision-4B model with its CoF-enhanced variant. Across all benchmarks, CoF-Phi-3.5-Vision-4B consistently surpasses the corresponding base model. The largest improvements appear on the \vsi benchmark, where CoF-Phi-3.5-Vision-4B achieves better performance compared to LLaVA-OneVision-7B and Qwen2-VL-7B (Tab.~\ref{tab:summary_results}) despite using substantially fewer parameters. These results demonstrate that our chain-of-frames approach generalizes effectively across architectures and model families.

% \paragraph{Comparison to SOTA.}
% \sara{Retrieval vs. referencing}
% We could not include the original models from other works \citep{wang2024videocot, han2024videoespresso, hao2024video, hu2025mllm} in the main evaluation in Table~\ref{tab:summary_results} because they are not accessible to the public, and they do not report results on the five benchmarks we use.
% However, we provide comparative analyses to M-LLM \citep{hu2025mllm} and Video-of-Thought \citep{hao2024video} on \videomme and \nextqa \citep{xiao2021next} using the results they report. 
% In Table~\ref{tab:additional_baseline}, we see that our CoF-based models, including CoF-\ivls which has fewer parameters than the competitors, outperform these baselines on both benchmarks.
% %
% Moreover, compared to the most recent model, M-LLM, our method achieves a greater improvement (4.9\% vs 0.8\%) on \nextqa despite starting from a stronger base model.

% \input{tables/nextqa}

\paragraph{Effect of CoT prompting.}
An extended version of Tab.~\ref{tab:reasoning_methods} is presented in Tab.~\ref{tab:prompt-analysis}.
For all baselines, we report results using two prompting strategies, either standard (indicated by $\boldsymbol{\star}$) or chain-of-thought (indicated by $\clubsuit$, the prompt is shown in Fig.~\ref{fig:cot-prompt}).
For our SFT with CoF models, we always use CoT prompting.
When considering \ivls, CoT prompting alone improves the accuracy of the original models on four out of five benchmarks compared to the original model.
However, this improvement does not hold for the SFT with QA only variant: we hypothesize that fine-tuning solely on QA data negatively impacts the reasoning capabilities of the baseline model. 
On the other hand, incorporating reasoning traces into the training data (SFT with CoT) generally enhances the model's reasoning capabilities, and using CoT prompting is beneficial except for the \eventhallusion benchmark. 
CoT prompting improves the results also for the original \ivlb on most benchmarks.
Finally, our models (SFT with CoF) outperform the baseline across all benchmarks.

\begin{table*}[t]
    \centering
    \small
    \tabcolsep=2pt
    %\vspace{2mm}
    %\resizebox{\textwidth}{!}{%
    \begin{tabular}{L{28mm}C{10mm}|*{6}{C{20mm}}}
    Model & Prompt &\textbf{\vsi} & \textbf{\videomme} & \textbf{\mvbench} &  \textbf{\vidhal} & \textbf{\eventhallusionshort}  \\
     \toprule 
    \multicolumn{3}{l}{\textbf{\ivls}} \\
     \midrule
     \multirow{2}{*}{Original}& $\boldsymbol{\star}$ & 31.8 & 54.9 & 70.8 & 74.0 & 62.5 \\
    & $\clubsuit$ & 33.5 & 54.7 & 71.5 & 77.0 & 67.4 \\
    \cmidrule{2-7}
    \multirow{2}{*}{SFT with QA only }& $\boldsymbol{\star}$ & 31.8 & 55.4 & 70.3 & 73.6 & 63.1 \\
    & $\clubsuit$ & 31.8 & 54.5 & 73.4 & 64.1 & 57.7 \\
    \cmidrule{2-7}
    \multirow{2}{*}{SFT with CoT} & $\boldsymbol{\star}$ & 31.1 & 52.6 & 69.6 & 74.4 & 62.5 \\ 
    & $\clubsuit$ & 34.3 & 58.6 & 73.7 & 77.9 & 53.1 \\
     \cmidrule{2-7}
     SFT with CoF (ours)  & $\clubsuit$ & 36.9 & 59.7 & 76.1 & 79.2 & 71.2 \\
    \midrule
    \multicolumn{3}{l}{\textbf{\ivlb}}\\
    \midrule    
    \multirow{2}{*}{Original} & $\boldsymbol{\star}$  & 41.0 & 62.3 & 72.0 & 80.9 & 72.1 \\
    & $\clubsuit$ & 40.2 & 66.5 & 74.3 & 61.6 & 73.9 \\
    \cmidrule{2-7}
     SFT with CoF (ours)  & $\clubsuit$ & 51.3 & 73.7 & 77.1 & 79.5 & 78.7 \\
    \midrule
     \multicolumn{3}{l}{\textbf{\phiv}}\\
     \midrule  
     \multirow{2}{*}{Original} & $\boldsymbol{\star}$ & 
     26.6 & 50.2 & 48.7 & 52.7 & 55.2\\
     & $\clubsuit$ & 
     26.4 & 39.0 & 46.6 & 43.1 & 51.3\\
    \cmidrule{2-7}
     SFT with CoF (ours) & $\clubsuit$ & 
     32.8 & 52.6 & 51.8 & 55.5 & 59.2 \\
    \bottomrule
    \end{tabular}%}
    \caption{\textbf{Effect of CoT prompting.} For all baselines, we report results using two prompting strategies, i.e. standard (indicated by $\boldsymbol{\star}$) and chain-of-thought (indicated by $\clubsuit$), while we fix CoT prompting for our CoF models.
    }
    \label{tab:prompt-analysis}
\end{table*}

\paragraph{Detailed results over benchmark splits.}
For completeness, we report the fine-grained results over the various splits of \vsi (Tab.~\ref{tab:vsi-details}), \videomme (Tab.~\ref{tab:mme-details}), \mvbench (Tab.~\ref{tab:mvbench-details}), and \eventhallusion (Tab.~\ref{tab:hallu-details}).
Moreover, for the baseline models, we report results using two prompting strategies, i.e., standard (indicated by $\boldsymbol{\star}$) and chain-of-thought (indicated by $\clubsuit$).

\begin{figure*}[t]
    \centering
     \includegraphics[width=.95\linewidth]{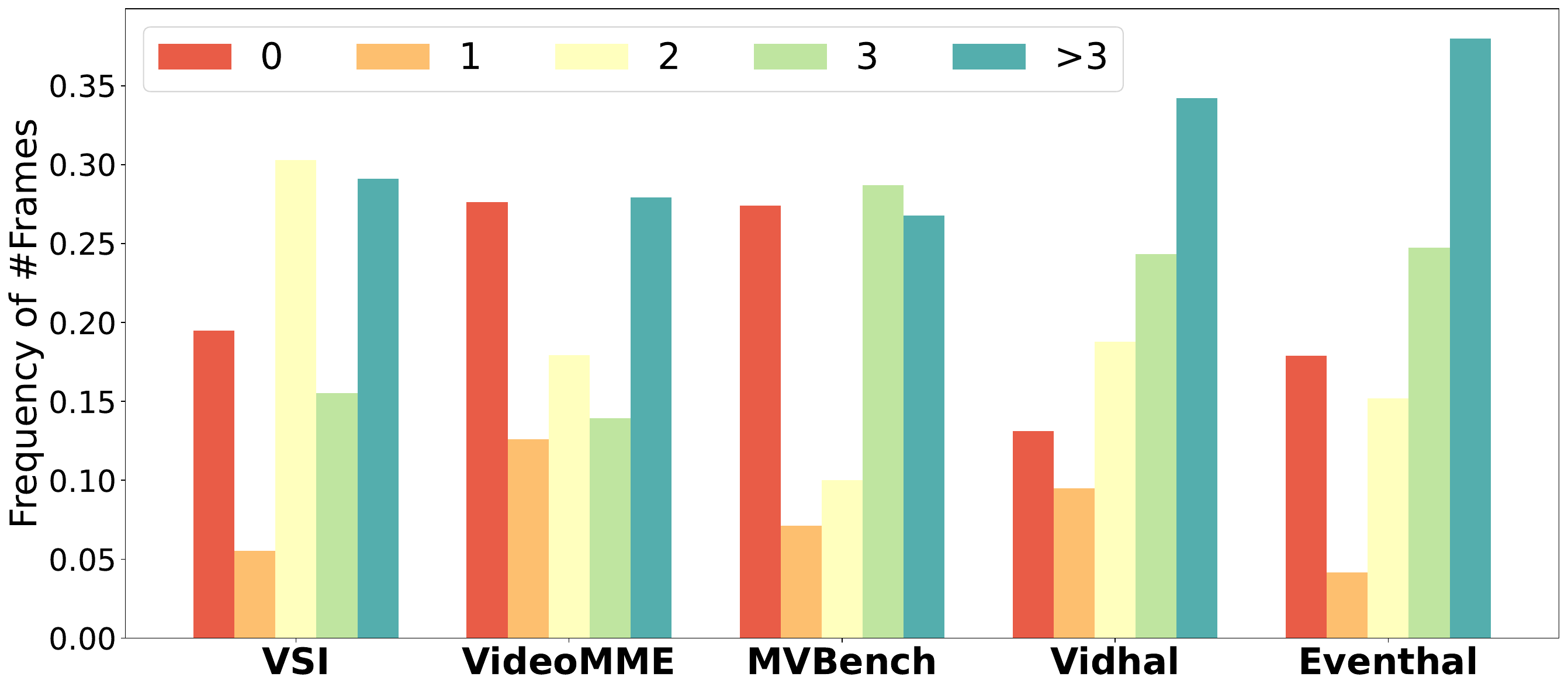}
    \caption{\textbf{CoF reasoning at inference time.} For each benchmark, we show the frequency of the number of frames referenced in the reasoning traces of \cofb.}
    \label{fig:frequency_frames_detailed}
\end{figure*}

%\subsection{Details on inference traces statistics.}
\paragraph{Detailed results of CoF reasoning at inference time.}
In Fig.~\ref{fig:frequency_frames_detailed}, we show statistics of how many frames are referenced in the reasoning traces generated by CoF-\ivlb.
To complement Fig.~\ref{fig:frame-analysis}, we report the frequency for each benchmark separately.
We see that the number of frames mentioned varies across benchmarks, e.g., the cases where no frames are referenced significantly decreases on the hallucination benchmarks \vidhal and \eventhallusion.
This suggests that our CoF models learn to modulate the reasoning traces and the frame references depending on the task.

\begin{table*}
\centering
\small
\extrarowheight=1.5pt
\tabcolsep=5pt

%\vspace{2mm}
%\resizebox{\textwidth}{!}{%
\begin{tabular}{L{28mm}C{10mm}|*{9}{C{8mm}}}

\textbf{Model} & Prompt & \rotatebox{70}{\textbf{Obj. Count}} & \rotatebox{70}{\textbf{Abs. Dist.}} & \rotatebox{70}{\textbf{Obj. Size}} & \rotatebox{70}{\textbf{Room Size}} & \rotatebox{70}{\textbf{Rel. Dist.}} & \rotatebox{70}{\textbf{App. Order}} & \rotatebox{70}{\textbf{Rel. Dir.}} & \rotatebox{70}{\textbf{Route Plan}} & \rotatebox{70}{\textbf{Avg}} \\
\toprule
    \multicolumn{3}{l}{\textbf{\ivls}} \\
     \midrule
     \multirow{2}{*}{Original}& $\boldsymbol{\star}$ & 29.2 & 31.2 & 45.5 & 20.4 & 35.4 & 23.2 & 41.4 & 27.8 & 31.8  \\
                              & $\clubsuit$ & 36.0 & 17.1 & 38.1 & 29.8 & 34.2 & 30.3 & 52.2 & 29.9 & 33.5 \\
    \cmidrule{2-11}
    \multirow{2}{*}{SFT with QA only }& $\boldsymbol{\star}$ & 22.7 & 30.7 & 44.0 & 26.0 & 36.3 & 22.2 & 39.4 & 33.0 & 31.8 \\
                                      & $\clubsuit$ & 34.9 & 18.6 & 38.8 & 23.2 & 37.0 & 28.2 & 47.1 & 26.3 & 31.8 \\
    \cmidrule{2-11}
    \multirow{2}{*}{SFT with CoT} & $\boldsymbol{\star}$ & 31.5 & 22.0 & 41.6 & 27.9 & 36.8 & 21.2 & 40.4 & 27.3 & 31.1\\ 
                                  & $\clubsuit$ & 39.1 & 19.5 & 36.1 & 26.9 & 36.1 & 30.1 & 57.1 & 29.4 & 34.3 \\
     \cmidrule{2-11}
     SFT with CoF (ours)  & $\clubsuit$ & 42.5 & 20.8 & 36.4 & 29.4 & 35.4 & 32.4 & 62.2 & 36.1 & 36.9 \\
    \midrule
    \multicolumn{3}{l}{\textbf{\ivlb}}\\
    \midrule    
    \multirow{2}{*}{Original} & $\boldsymbol{\star}$ & 58.6 & 28.5 & 49.5 & 43.3 & 47.0 & 38.5 & 31.3 & 31.4 & 41.0 \\
                             & $\clubsuit$ & 55.2 & 29.5 & 38.1 & 32.6 & 42.8 & 47.1 & 47.5 & 28.4 & 40.2 \\
    \cmidrule{2-11}
     SFT with CoF (ours)  & $\clubsuit$ & 61.8 & 34.2 & 37.7 & 26.7 & 66.8 & 43.4 & 83.9 & 55.7 & 51.3 \\
     \midrule
     \multicolumn{3}{l}{\textbf{\phiv}}\\
     \midrule  
     \multirow{2}{*}{Original} & $\boldsymbol{\star}$ & 27.3 &	23.4 & 30.3 & 21.4 & 26.5 & 17.5 & 33.7 & 33.0 & 26.6 \\
     & $\clubsuit$ & 27.3 & 24.1 & 32.6 & 22.6 & 20.5 & 20.9	& 35.2 & 28.3 & 26.4 \\
    \cmidrule{2-11}
     SFT with CoF (ours)  & $\clubsuit$ & 31.5 & 27.9 & 37.9 & 26.3 & 33.5 & 22.7 & 47.3 & 35.1 & 32.8  \\
     \bottomrule
\end{tabular}%}
\caption{\textbf{Detailed results on the \vsi benchmark.} For the baselines, we report results using two prompting strategies, i.e. standard (indicated by $\boldsymbol{\star}$) and chain-of-thought (indicated by $\clubsuit$), while we fix chain-of-thoughts prompting for our CoF models.}
\label{tab:vsi-details}
\end{table*}

\begin{table*}[p]
\centering
\small
\extrarowheight=1.5pt

%\vspace{2mm}
%\resizebox{\textwidth}{!}{%
\begin{tabular}{L{28mm}C{10mm}|*{3}{C{20mm}}C{10mm}}
\textbf{Model} & Prompt & \textbf{Short (900)} & \textbf{Medium (900)} & \textbf{Long (900)} & \textbf{Avg} \\
\toprule
    \multicolumn{3}{l}{\textbf{\ivls}} \\
     \midrule
     \multirow{2}{*}{Original}& $\boldsymbol{\star}$ &  64.9 & 52.7 & 47.2 & 54.9 \\
                              & $\clubsuit$ & 64.0 & 53.2 & 47.0 & 54.7 \\
    \cmidrule{2-6}
    \multirow{2}{*}{SFT with QA only }& $\boldsymbol{\star}$ & 68.0 & 53.6 & 44.8 & 55.5 \\
                                      & $\clubsuit$ & 66.8 & 53.1 & 43.6 & 54.5 \\
    \cmidrule{2-6}
    \multirow{2}{*}{SFT with CoT} & $\boldsymbol{\star}$ & 64.3 & 51.8 & 41.8 & 52.6 \\
                                  & $\clubsuit$ & 70.4 & 55.7 & 49.6 & 58.6 \\
     \cmidrule{2-6}
     SFT with CoF (ours)  & $\clubsuit$ & 73.1 & 56.2 & 49.9 & 59.7 \\
    \midrule
    \multicolumn{3}{l}{\textbf{\ivlb}}\\
    \midrule    
    \multirow{2}{*}{Original} & $\boldsymbol{\star}$ & 73.0 & 61.7 & 52.1 & 62.3 \\
     & $\clubsuit$ &  75.3 & 65.3 & 59.0 & 66.6 \\
    \cmidrule{2-6}
     SFT with CoF (ours)  & $\clubsuit$ &  79.3 & 71.7 & 70.0 & 	73.7 \\
     \midrule
     \multicolumn{3}{l}{\textbf{\phiv}}\\
     \midrule  
     \multirow{2}{*}{Original} & $\boldsymbol{\star}$ & 61.4 & 50.6 & 38.7 & 50.2\\
     & $\clubsuit$ & 46.0 & 34.9 & 36.1 & 39.0 \\
    \cmidrule{2-6}
     SFT with CoF (ours)  & $\clubsuit$ & 63.1 & 51.9 & 42.7 & 52.6 \\
     \bottomrule
\end{tabular}%}
\caption{\textbf{Detailed results on the \videomme benchmark.} For the baselines, we report results using two prompting strategies, i.e. standard (indicated by $\boldsymbol{\star}$) and chain-of-thought (indicated by $\clubsuit$), while we fix chain-of-thoughts prompting for our CoF models.}
\label{tab:mme-details}
\end{table*}

\begin{table*}[p]
\centering
\tabcolsep=4pt
%\vspace{2mm}
%\resizebox{\linewidth}{!}{
\begin{tabular}{L{30mm}C{10mm}|*{10}{C{9mm}}}

\textbf{Model} & Prompt & \textbf{AA} & \textbf{AC} & \textbf{AL} & \textbf{AP} & \textbf{AS} & \textbf{CO} & \textbf{CI} & \textbf{EN} & \textbf{FA} & \textbf{MA} \\
\toprule
    \multicolumn{3}{l}{\textbf{\ivls}} \\
     \midrule
     \multirow{2}{*}{Original}& $\boldsymbol{\star}$ &  89.5 & 54.0 & 44.0 & 75.0 & 82.9 & 62.5 & 79.0 & 29.0 & 46.0 & 97.5 \\
                              & $\clubsuit$ & 88.5 & 50.0 & 46.5 & 77.0 & 81.4 & 67.0 & 78.0 & 34.5 & 60.5 & 99.0 \\
    \cmidrule{2-12}
    \multirow{2}{*}{SFT with QA only} & $\boldsymbol{\star}$ & 90.0 & 55.0 & 44.0 & 76.5 & 81.9 & 63.5 & 75.0 & 33.0 & 46.0 & 98.5 \\
                                      & $\clubsuit$ &  90.5 & 50.0 & 48.5 & 78.5 & 84.0 & 67.5 & 80.0 & 38.5 & 71.5 & 99.5 \\
    \cmidrule{2-12}
    \multirow{2}{*}{SFT with CoT} & $\boldsymbol{\star}$ & 87.0 & 53.0 & 35.5 & 76.5 & 81.4 & 62.0 & 76.5 & 31.5 & 43.5 & 98.0 \\
                                  & $\clubsuit$ & 90.5 & 46.5 & 57.5 & 85.0 & 83.5 & 67.0 & 80.5 & 38.5 & 73.5 & 98.5 \\
     \cmidrule{2-12}
     SFT with CoF (ours)  & $\clubsuit$ & 93.0 & 41.0 & 62.0 & 91.5 & 89.4 & 73.5 & 79.5 & 47.0 & 83.0 & 98.5 \\
    \midrule
    \multicolumn{3}{l}{\textbf{\ivlb}}\\
    \midrule    
    \multirow{2}{*}{Original} & $\boldsymbol{\star}$ & 
    90.0 & 42.0 & 44.5 & 83.0 & 82.45 & 75.5 & 78.0 & 38.5 & 45.0 & 98.0\\
    & $\clubsuit$ & 77.5 & 45.0 & 42.5 & 87.5 & 85.1 & 80.5 & 89.0 & 33.5 & 47.0	& 99.0  \\
    \cmidrule{2-12}
     SFT with CoF (ours)  & $\clubsuit$ & 96.5 & 50.5 & 49.5 & 89.5 & 91.0 & 91.5 & 77.5 & 45.5 & 59.5 & 	96.5   \\
    \midrule
     \multicolumn{3}{l}{\textbf{\phiv}}\\
     \midrule  
     \multirow{2}{*}{Original} & $\boldsymbol{\star}$ & 
65.0 & 52.5 & 32.0 & 45.5 & 40.96 & 48.5 & 36.5 & 40.0 & 36.0 & 65.5 \\
     & $\clubsuit$ & 61.0 & 49.5 & 27.5 & 43.0 & 37.8 & 42.0 & 35.0 & 36.5 & 36.5  & 61.5 \\
    \cmidrule{2-12}
     SFT with CoF (ours)  & $\clubsuit$ & 	
68.0 & 51.5 & 38.5 & 49.5 & 42.02 & 49.5 & 39.5 & 44.0 & 38.5 & 68.0 \\
     \bottomrule

\addlinespace[10mm]

\textbf{Model} & Prompt & &\textbf{MC} & \textbf{MD} & \textbf{OE} & \textbf{OI} & \textbf{OS} & \textbf{ST} & \textbf{SC} & \textbf{UA} & \textbf{Avg} \\
\toprule
    \multicolumn{3}{l}{\textbf{\ivls}} \\
     \midrule
     \multirow{2}{*}{Original}& $\boldsymbol{\star}$ &  &88.5 & 73.0 & 96.5 & 83.5 & 39.5 & 92.0 & 57.5 & 85.0 & 70.8 \\
                              & $\clubsuit$ &  &86.5 & 72.5 & 96.0 & 81.5 & 40.5 & 91.5 & 58.0 & 78.0 & 71.5 \\
    \cmidrule{2-12}
    \multirow{2}{*}{SFT with QA only }& $\boldsymbol{\star}$ & & 87.5 & 75.0 & 96.5 & 82.5 & 41.0 & 91.5 & 59.5 & 85.5 & 70.3 \\
                                      & $\clubsuit$ & & 86.5 & 75.5 & 96.5 & 86.5 & 42.0 & 92.0 & 52.5 & 82.0 & 73.4 \\
    \cmidrule{2-12}
    \multirow{2}{*}{SFT with CoT} & $\boldsymbol{\star}$ & & 89.0 & 72.5 & 96.5 & 82.0 & 38.0 & 91.5 & 56.5 & 82.0 & 69.6 \\
                                  & $\clubsuit$ & &86.5 & 72.0 & 95.5 & 86.5 & 40.5 & 92.0 & 52.5 & 80.5 & 73.7 \\
     \cmidrule{2-12}
     SFT with CoF (ours)  & $\clubsuit$ & & 86.5 & 72.0 & 96.5 & 87.0 & 44.0 & 93.5 & 50.0 & 82.5 & 76.1 \\
    \midrule
    \multicolumn{3}{l}{\textbf{\ivlb}}\\
    \midrule    
    \multirow{2}{*}{Original} & $\boldsymbol{\star}$ & & 60.5 & 89.0 & 96.97 & 85.5 & 39.5 & 92.5 & 69.0 & 81.0 & 71.7 \\
    & $\clubsuit$ && 63.5 & 89.0 & 97.47 & 87.5 & 41.5 & 92.5 & 73.0 & 73.5 & 72.5\\
    \cmidrule{2-12}
     SFT with CoF (ours)  & $\clubsuit$ && 67.0 & 91.0 & 90.4 & 90.5 & 45.5 & 94.0 & 78.0 & 84.0 & 77.1 \\
     \midrule
     \multicolumn{3}{l}{\textbf{\phiv}}\\
     \midrule  
     \multirow{2}{*}{Original} & $\boldsymbol{\star}$ & & 35.0 & 37.0 & 57.1 & 45.0 & 31.0 & 88.5 & 46.0 & 75.0 & 48.7\\
     & $\clubsuit$ & & 35.5 & 38.5 & 53.0 & 43.5 & 28.5 & 89.0 & 47.5 & 72.5 & 46.6 \\
    \cmidrule{2-12}
     SFT with CoF (ours)  & $\clubsuit$ && 36.5 & 45.5 & 61.1 & 46.5 & 33.5 & 90.0 & 53.3 & 76.5 & 51.8  \\
     \bottomrule
\end{tabular}%}
\caption{\textbf{Detailed results on the \mvbench benchmark.} For the baselines, we report results using two prompting strategies, i.e. standard (indicated by $\boldsymbol{\star}$) and chain-of-thought (indicated by $\clubsuit$), while we fix chain-of-thoughts prompting for our CoF models.}
\label{tab:mvbench-details}
\end{table*}

\begin{table*}[p]
\centering
\small
\extrarowheight=1.5pt

%\vspace{2mm}
% \resizebox{\linewidth}{!}{%
\begin{tabular}{L{28mm}C{10mm}|C{10mm}@{\hspace{0.5cm}}*{4}{C{12mm}}}
\multirow{2}{*}{\textbf{Model}} & \multirow{2}{*}{Prompt} & \multirow{2}{*}{\textbf{\vidhal}} & \multicolumn{4}{c}{\textbf{\eventhallusion}} \\
\cmidrule{4-7}
& & & Entire & Misleading & Mix & Avg \\
\toprule
    \multicolumn{3}{l}{\textbf{\ivls}} \\
     \midrule
     \multirow{2}{*}{Original}& $\boldsymbol{\star}$ & 74.0 & 48.3 & 91.2 & 48.2 & 62.5 \\
                              & $\clubsuit$ &  77.0 & 44.7 & 80.4 & 77.2 & 67.4 \\
    \cmidrule{2-7}
    \multirow{2}{*}{SFT with QA only }& $\boldsymbol{\star}$ & 73.6 & 48.3 & 89.2 & 51.8 & 63.1 \\
                                      & $\clubsuit$ & 64.1 & 47.4 & 75.5 & 50.3 & 57.7 \\
    \cmidrule{2-7}
    \multirow{2}{*}{SFT with CoT} & $\boldsymbol{\star}$ & 74.4 & 49.1 & 91.2 & 47.1 & 62.5 \\
                                  & $\clubsuit$ & 77.9 & 39.5 & 71.6 & 48.2 & 53.1 \\
     \cmidrule{2-7}
     SFT with CoF (ours)  & $\clubsuit$ & 79.2 & 49.1 & 85.3 & 79.3 & 71.2 \\
    \midrule
    \multicolumn{3}{l}{\textbf{\ivlb}}\\
    \midrule    
    \multirow{2}{*}{Original} & $\boldsymbol{\star}$ & 80.9 & 52.6 & 91.2 & 72.5 & 72.1 \\
                             & $\clubsuit$ & 61.6 & 57.0 & 94.1 & 70.5 & 73.9 \\
    \cmidrule{2-7}
     SFT with CoF (ours)  & $\clubsuit$ & 79.5 & 57.9 & 92.2 & 86.0 & 78.7 \\
    \midrule
     \multicolumn{3}{l}{\textbf{\phiv}}\\
     \midrule  
     \multirow{2}{*}{Original} & $\boldsymbol{\star}$ & 52.7 & 27.2 & 72.6 & 65.8 & 55.2 \\
     & $\clubsuit$ & 43.1 & 22.8 & 64.7 & 66.3 & 51.3 \\
    \cmidrule{2-7}
     SFT with CoF (ours)  & $\clubsuit$ & 55.5 & 34.2 & 73.5 & 70.0 & 59.2 \\
     \bottomrule
\end{tabular}
\caption{\textbf{Detailed results on the \vidhal and \eventhallusion benchmarks.} For the baselines, we report results using two prompting strategies, i.e. standard (indicated by $\boldsymbol{\star}$) and chain-of-thought (indicated by $\clubsuit$), while we fix chain-of-thoughts prompting for our CoF models.}
\label{tab:hallu-details}
\end{table*}

\section{Additional Figures}
\label{sec:additional_figures}

\begin{figure*}[h]
    \centering
    \includegraphics[width=\linewidth, trim=0mm 55mm 0mm 0mm, clip]{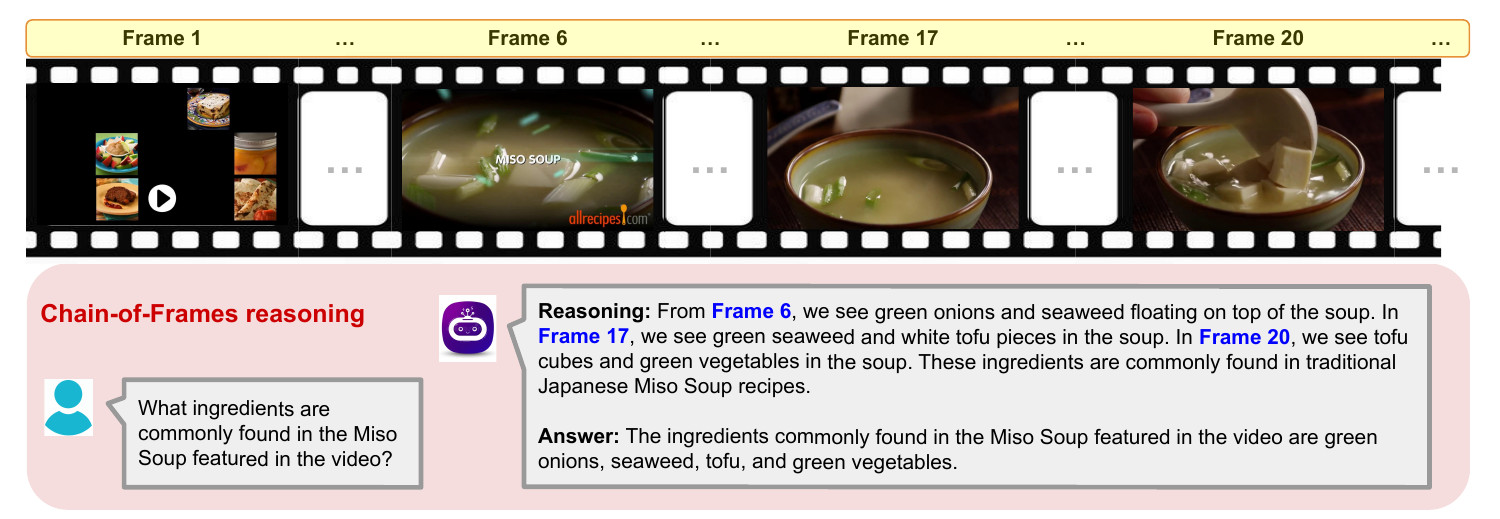}
    \includegraphics[width=\linewidth, trim=0mm 50mm 0mm 0mm, clip]{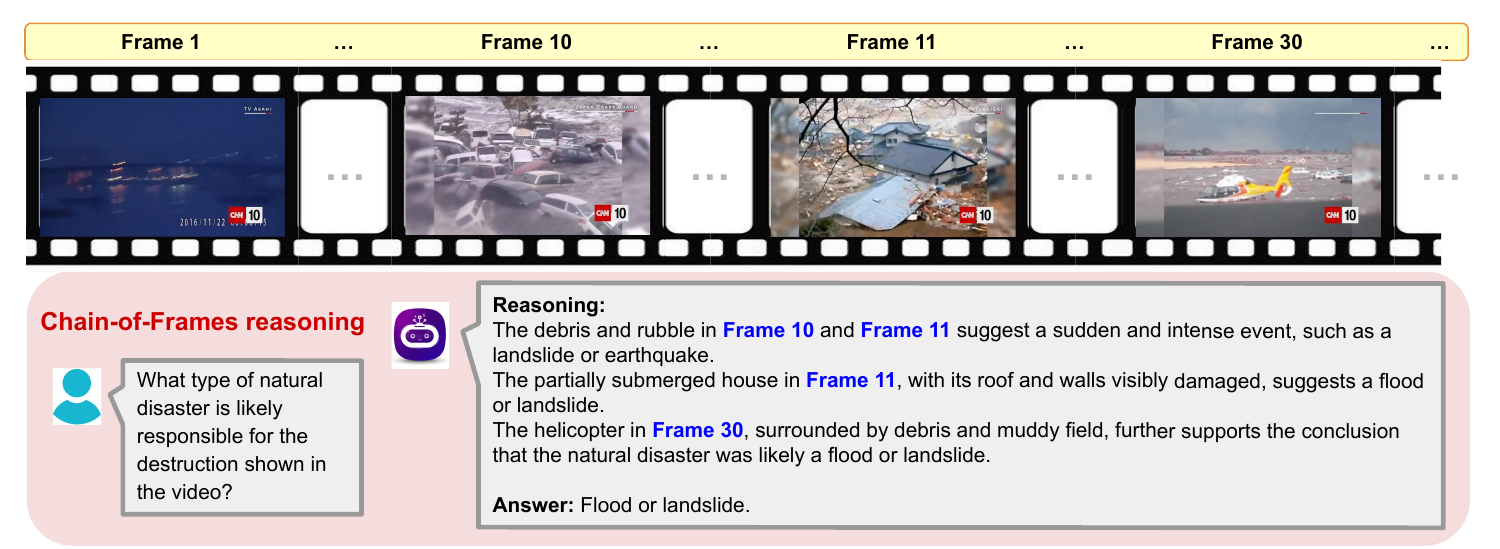}
    \caption{\textbf{CoF training data.} Examples of the training data from \cofdatareal, including relevant frames, questions, answers, and reasoning traces.}
    \label{fig:cof_videoespresso}
\end{figure*}

\begin{figure*}
    \centering
    \includegraphics[width=\linewidth, trim=10mm 55mm 10mm 0mm, clip]{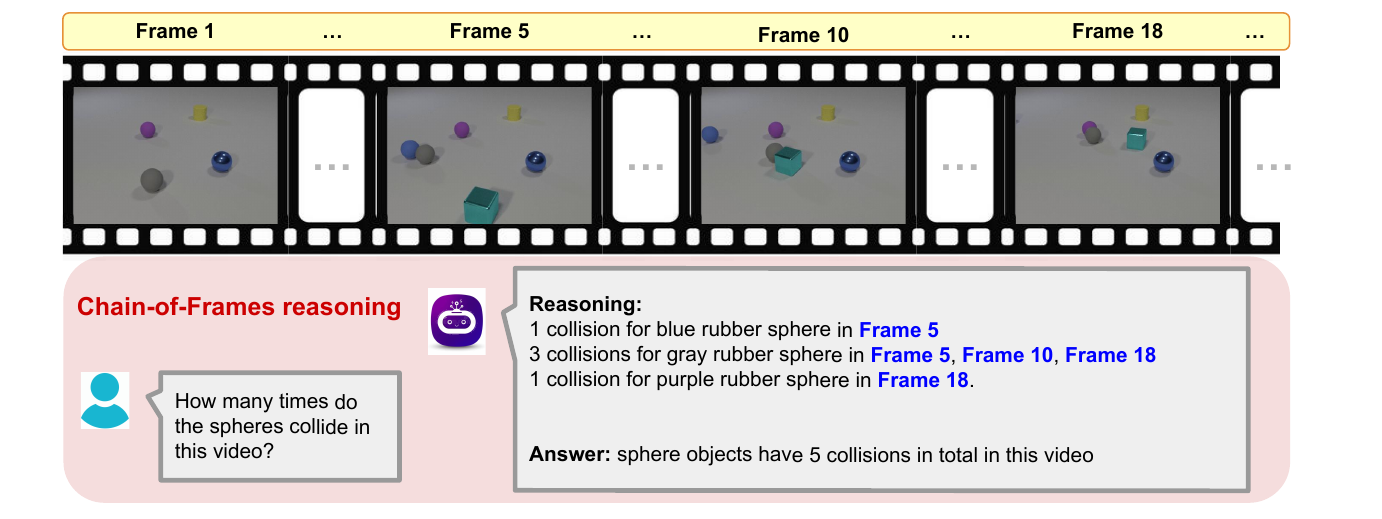}
    \includegraphics[width=\linewidth, trim=10mm 25mm 10mm 0mm, clip]{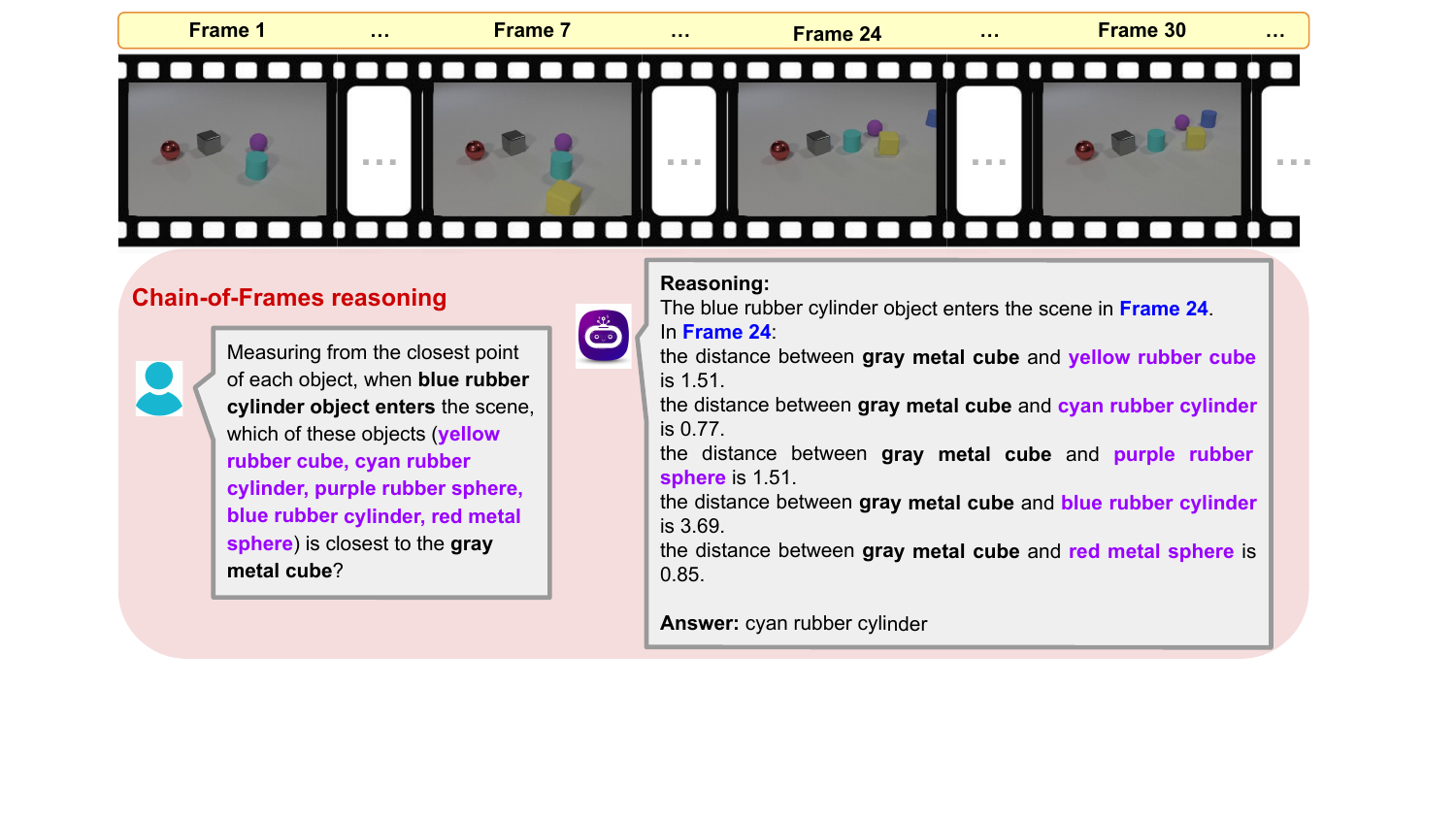}
    \caption{\textbf{CoF training data.} Examples of training data generated from the \cofdatasyn dataset, including relevant frames, questions, answers, and reasoning traces. The samples shown belong to the \textit{object count} and \textit{relative distance} categories, respectively.}
    \label{fig:cof_clevrer}
\end{figure*}

\begin{figure*}
    \centering
     \begin{subfigure}{\linewidth}
    \includegraphics[width=\linewidth, trim=0mm 0mm 0mm 0mm, clip]{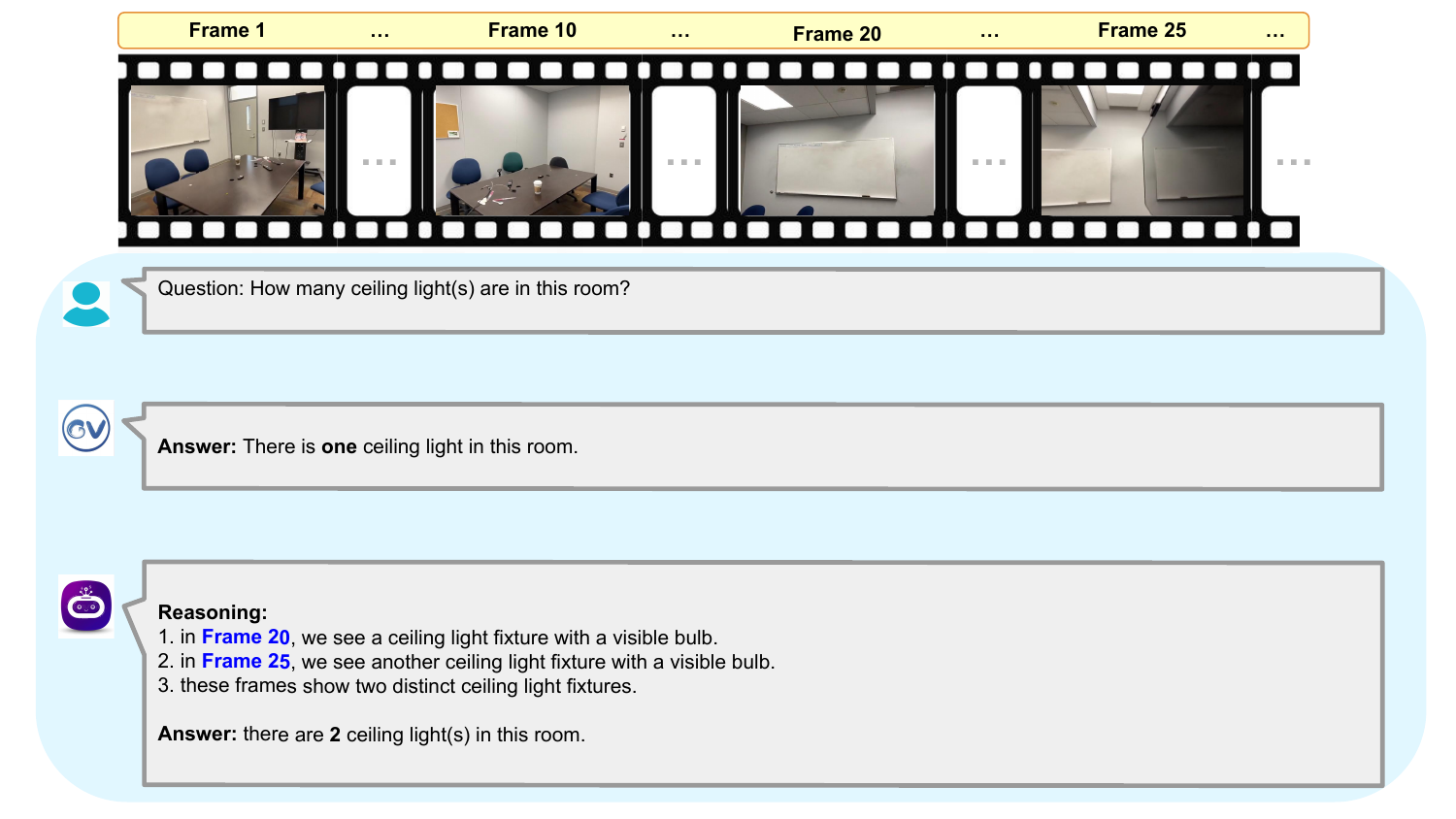} 
    \caption{\textbf{\vsi benchmark.} We show the question (first box), the answer and possibly CoT reasoning of the original \ivls with CoT prompting (second box), and the answer with CoF reasoning of our CoF-\ivls model (third box).}
    \end{subfigure}
    %\vspace{5mm}
    \begin{subfigure}{\linewidth}
    \includegraphics[width=\linewidth, trim=0mm 0mm 0mm 0mm, clip]{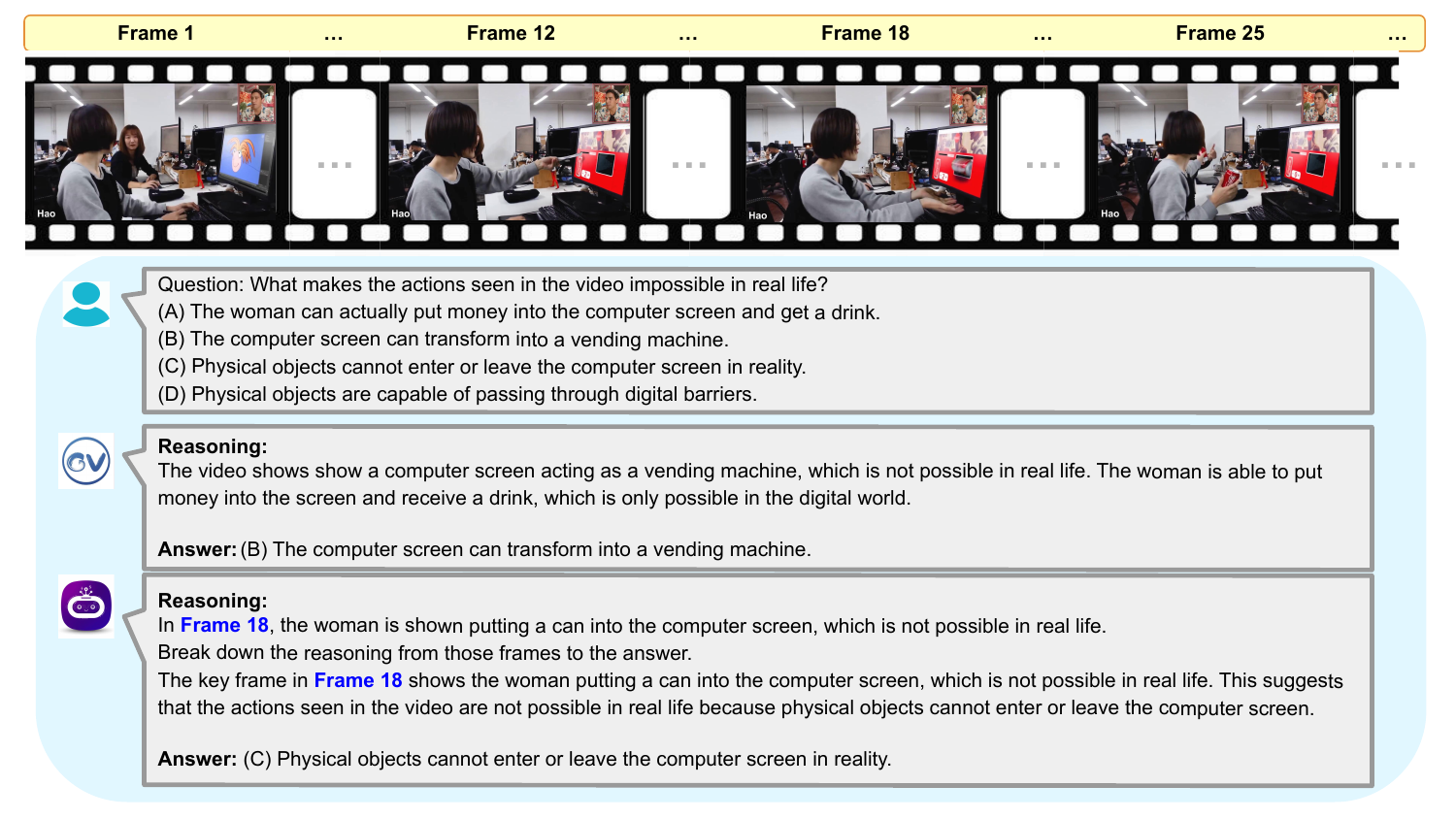} 
    \caption{\textbf{\mvbench benchmark.} We show the question (first box), the answer and possibly CoT reasoning of the original \ivlb with CoT prompting (second box), and the answer with CoF reasoning of our CoF-\ivlb model (third box).}
    \end{subfigure}
    \caption{\textbf{Inference examples.}}
    \label{fig:inference_examples}
\end{figure*}

\begin{figure*}
    \centering
    \begin{subfigure}{\linewidth}
    \includegraphics[width=\linewidth, trim=0mm 0mm 0mm 0mm, clip]{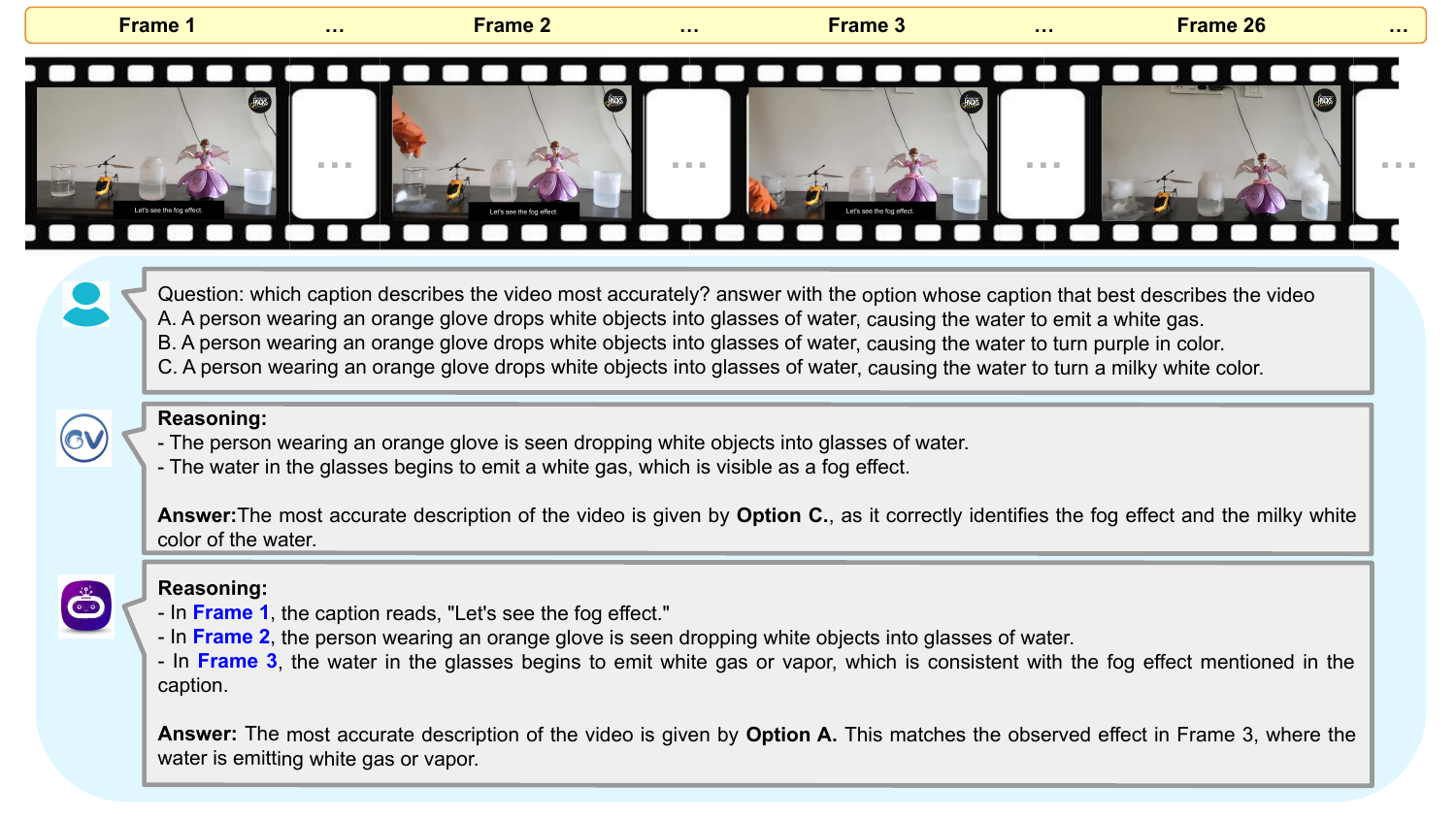}
    \caption{\textbf{\vidhal benchmark.} We show the question (first box), the answer, and possibly CoT reasoning of the original \ivls with CoT prompting (second box), and the answer with CoF reasoning of our CoF-\ivls model (third box).}
    \end{subfigure}
    \begin{subfigure}{\linewidth}
    \includegraphics[width=\linewidth, trim=0mm 20mm 0mm 0mm, clip]{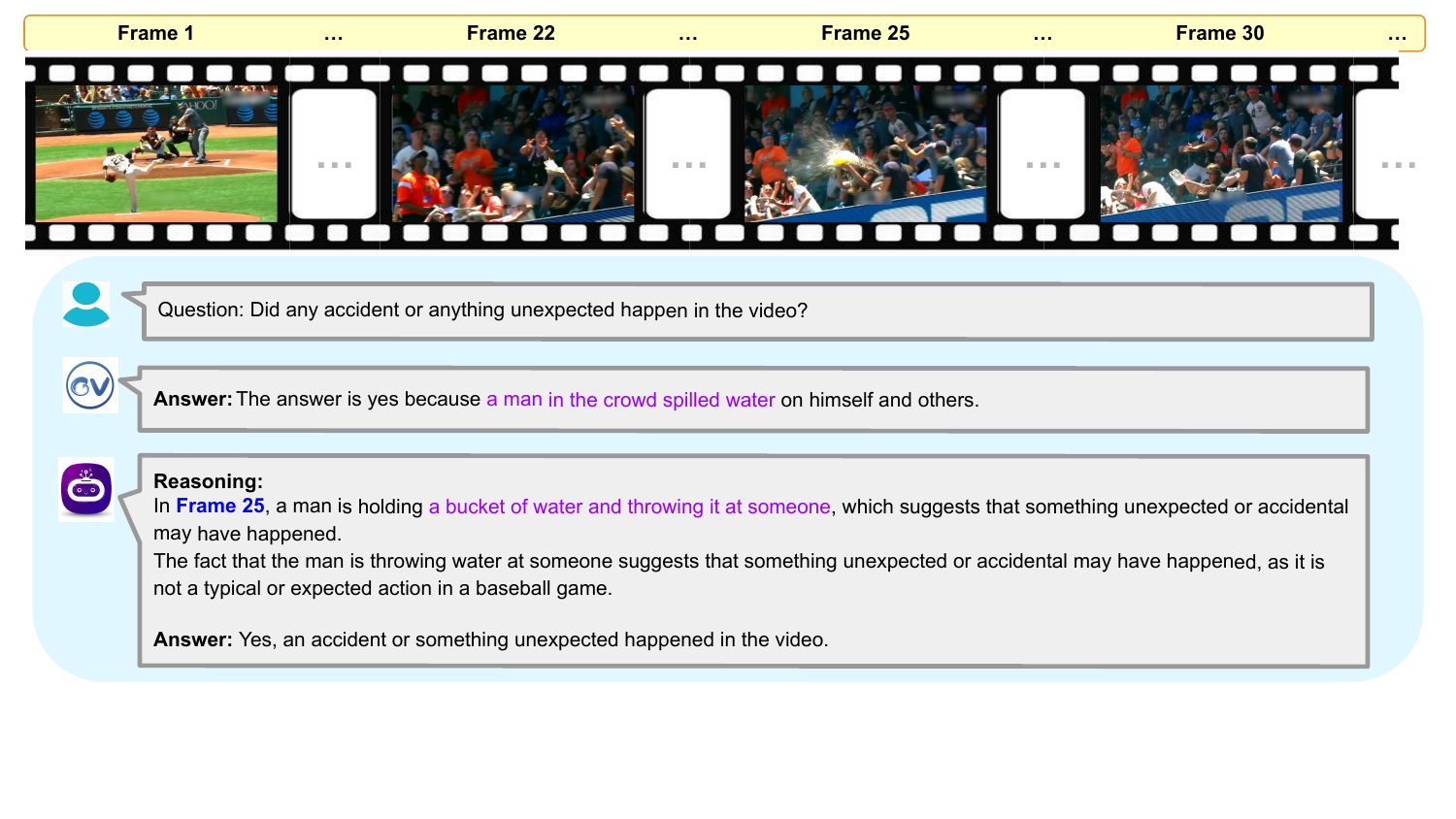}
    \caption{\textbf{\eventhallusion benchmark.} We show the question (first box), the answer, and possibly CoT reasoning of the original \ivlb with CoT prompting (second box), and the answer with CoF reasoning of our CoF-\ivlb model (third box).}
    \end{subfigure}
    \caption{\textbf{Inference examples.}}
    \label{fig:inference_hallu_examples}
\end{figure*}

This section presents additional samples from our training dataset along with inference examples. More specifically, Fig.~\ref{fig:cof_videoespresso} and Fig.~\ref{fig:cof_clevrer} show samples from the \cofdatareal and \cofdatasyn, respectively. 
To illustrate the reasoning traces generated by our CoF models and compare them to the answers of the baseline models, we present samples from \vsi and \mvbench benchmarks in Fig.~\ref{fig:inference_examples} and samples from hallucination benchmarks in Fig.~\ref{fig:inference_hallu_examples}.


\end{document}